\documentclass[journal]{IEEEtran}
\usepackage{cite}
\ifCLASSINFOpdf
\else
\fi
\usepackage{algorithmic}
\ifCLASSOPTIONcompsoc
 \usepackage[caption=false,font=normalsize,labelfont=sf,textfont=sf]{subfig}
\else
 \usepackage[caption=false,font=footnotesize]{subfig}
\fi

\usepackage{hyperref}
\usepackage{amsmath,amssymb,amsfonts}
\usepackage{amssymb}
\usepackage{breqn}
\usepackage{cleveref}
\usepackage[lined,ruled,linesnumbered,commentsnumbered]{algorithm2e}
\usepackage{booktabs}
\usepackage{cases}
\usepackage{graphicx}
\usepackage{multirow}
\usepackage{mathrsfs}
\usepackage{textcomp}
\usepackage{xcolor}
\usepackage{boxedminipage}
\usepackage{mathrsfs}
\usepackage{enumerate}
\usepackage{colortbl}
\usepackage{array}

\usepackage{graphbox}

\definecolor{DeepRed}{RGB}{202,49,66}

\begin{document}

\title{Improving Performance Insensitivity of Large-scale Multiobjective Optimization via Monte Carlo Tree Search}

\author{
  Haokai~Hong, ~\IEEEmembership{Graduate Student Member, ~IEEE,}
  Min~Jiang, ~\IEEEmembership{Senior Member,~IEEE,}
  and~Gary~G.~Yen, ~\IEEEmembership{Fellow,~IEEE}
  
  \thanks{H. Hong and M. Jiang are with the Department of Artificial Intelligence, Key Laboratory of Digital Protection and Intelligent Processing of Intangible Cultural Heritage of Fujian and Taiwan, Ministry of Culture and Tourism, School of Informatics, Xiamen University, Fujian, China, 361005.} %
  \thanks{G.G. YEN is with the School of Electrical and Computer Engineering, Oklahoma State University, USA. }
  \thanks{The corresponding author: Min Jiang, minjiang@xmu.edu.cn.}
}

\markboth{IEEE Transactions on Cybernetics,~Vol.~, No.~, ~}%
{Shell \MakeLowercase{\textit{et al.}}: A Sample Article Using IEEEtran.cls for IEEE Journals}


\maketitle

\begin{abstract}
The large-scale multiobjective optimization problem (LSMOP) is characterized by simultaneously optimizing multiple conflicting objectives and involving hundreds of decision variables. {Many real-world applications in engineering fields can be modeled as LSMOPs; simultaneously, engineering applications require insensitivity in performance.} This requirement usually means that the results from the algorithm runs should not only be good for every run in terms of performance but also that the performance of multiple runs should not fluctuate too much, i.e., the algorithm shows good insensitivity. Considering that substantial computational resources are requested for each run, it is essential to improve upon the performance of the large-scale multiobjective optimization algorithm, as well as the insensitivity of the algorithm. However, existing large-scale multiobjective optimization algorithms solely focus on improving the performance of the algorithms, leaving the insensitivity characteristics unattended. {In this work, we propose an evolutionary algorithm for solving LSMOPs based on Monte Carlo tree search, the so-called LMMOCTS, which aims to improve the performance and insensitivity for large-scale multiobjective optimization problems.} The proposed method samples the decision variables to construct new nodes on the Monte Carlo tree for optimization and evaluation. {It selects nodes with good evaluation for further search to reduce the performance sensitivity caused by large-scale decision variables.} We compare the proposed algorithm with several state-of-the-art designs on different benchmark functions. We also propose two metrics to measure the sensitivity of the algorithm. The experimental results confirm the effectiveness and performance insensitivity of the proposed design for solving large-scale multiobjective optimization problems.

\end{abstract}

\begin{IEEEkeywords}
Evolutionary multiobjective optimization, large-scale optimization, Monte Carlo Tree Search, Insensitivity. 
\end{IEEEkeywords}

\section{Introduction}
Optimization problems involving multiple conflicting objectives with more than 100 decision variables are known as large-scale multiobjective optimization problems (LSMOPs) \cite{7155533, 8681243}. {A variety of real-world applications can be modeled as LSMOPs \cite{8315121, 7533424, 8482477, 10.1007/978-3-319-16549-3_56, 8781874, 9804338}.} A good example in channel routing is the configuration of a wireless network with thousands of routers \cite{iqbal2015wireless}, which could be modeled as an optimization problem with thousands of decision variables and the concerned objectives in the energy consumption, network stability, QoS, and user experience. {Since the increase in the number of decision variables naturally leads to explosion of the search space in LSMOPs \cite{RN105},} finding the set of Pareto optimal solutions in LSMOPs is considered more challenging than ordinary multiobjective optimization problems (MOPs) \cite{5415586}.
\par
In recent years, numerous large-scale multiobjective optimization evolutionary algorithms (MOEAs) have been proposed. In \cite{6557903}, the authors proposed a method called CCGDE3, which divides the decision variables into several subgroups and then optimizes each subgroup in turn. Zille \emph{et al.} \cite{RN89} proposed a weighted optimization framework (WOF) for solving LSMOPs, in which conventional MOEAs are adopted to optimize the weight vector for the optimal objective values of the weighted solution on the LSMOPs. {The third representative large-scale MOEA is S3-CMA-ES proposed by Chen \emph{et al.} \cite{CHEN2020457}, which uses a scalable small subpopulation-based covariance matrix adaptation evolution strategy to solve the LSMOP.}
\par
Most of them, if not all, of these methods focus solely on dealing with the massive search space \cite{8720021, 9047876, 9804338}. The main reason is that the search space for optimization problems will increase exponentially with the increase in decision variables. {In our cited example, the large number of routers considered results in an exponentially growing search space.} Appreciable progress has been made in addressing LSMOPs, yet existing designs are far from applicable to cope with real-world complications. 
\par
\begin{figure*}[htbp]
    \centering
    \subfloat[{CCGDE3 Solves  LSMOP3}]{\includegraphics[width=0.24\hsize]{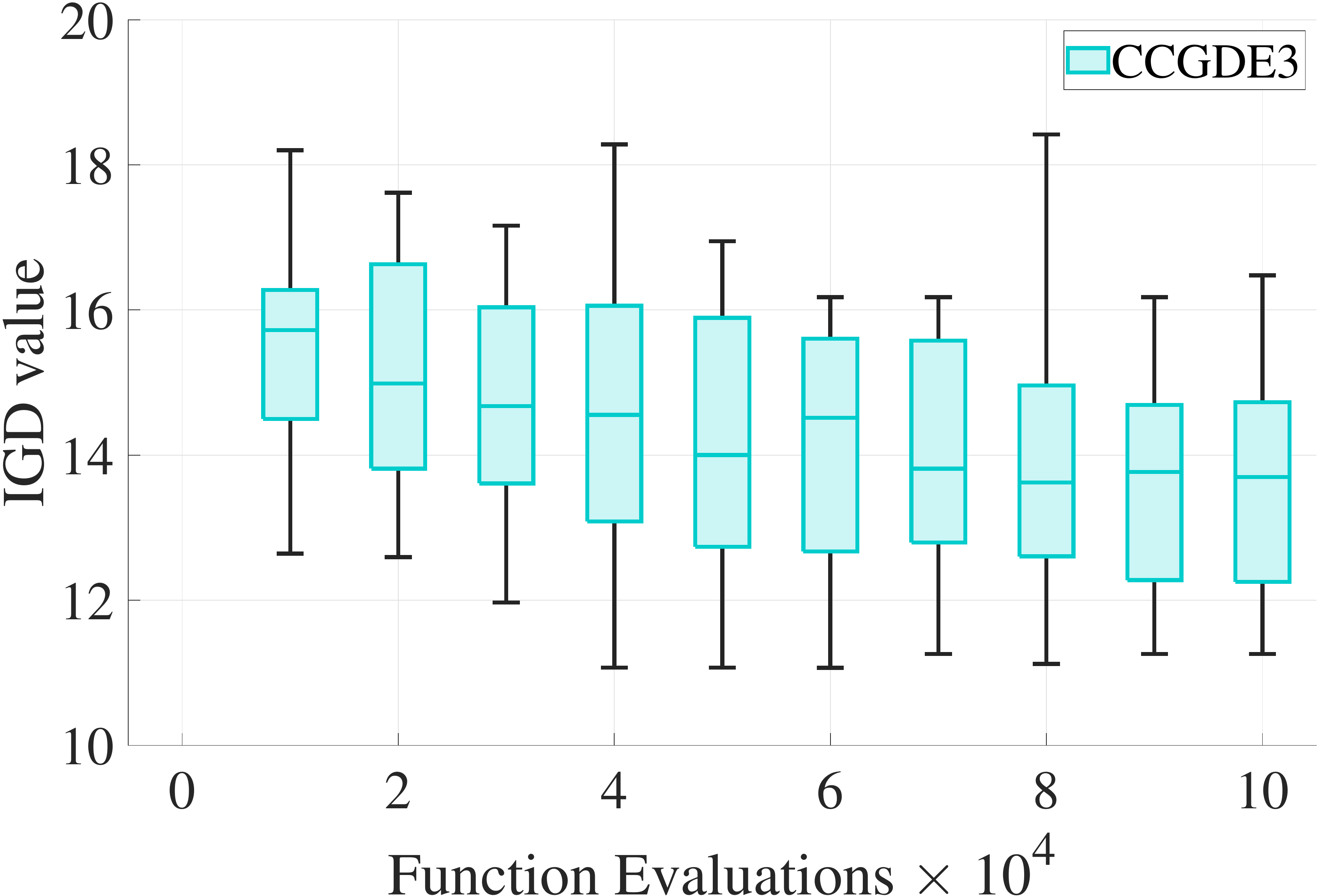}}
    \subfloat[{CCGDE3 Solves  LSMOP6}]{\includegraphics[width=0.24\hsize]{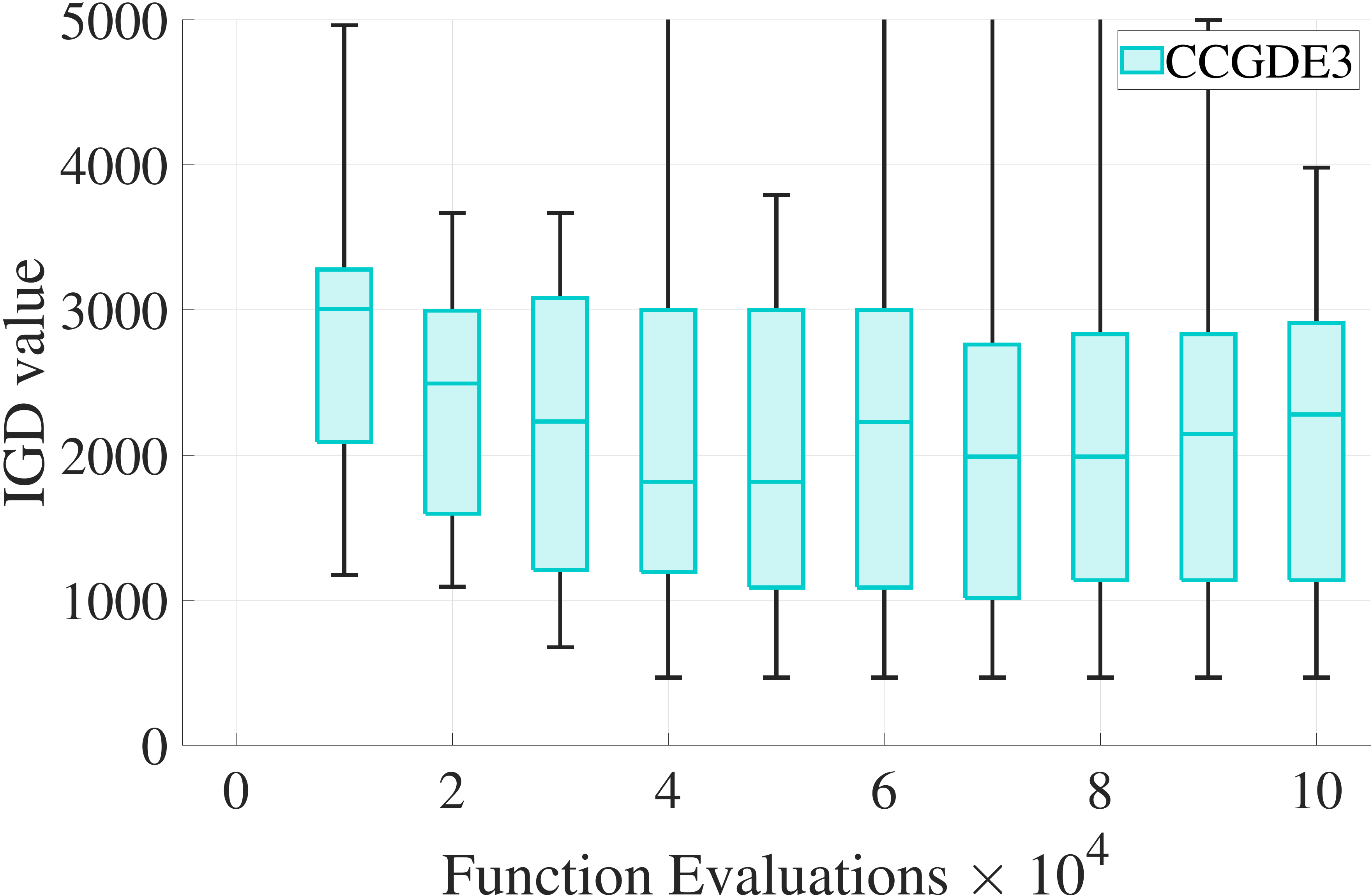}}
    \subfloat[{LMOCSO Solves  LSMOP3}]{\includegraphics[width=0.24\hsize]{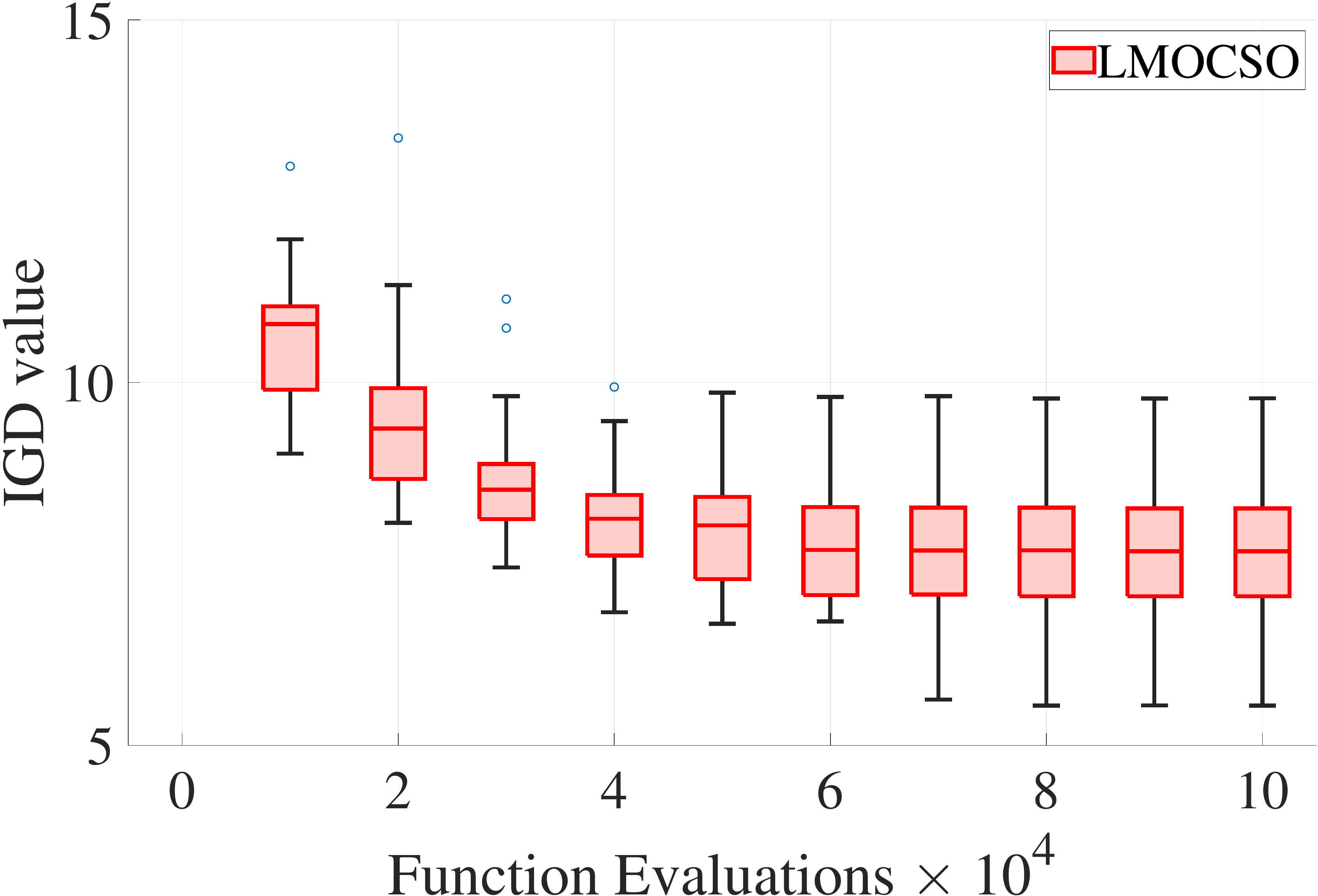}}
    \subfloat[{LMOCSO Solves  LSMOP6}]{\includegraphics[width=0.24\hsize]{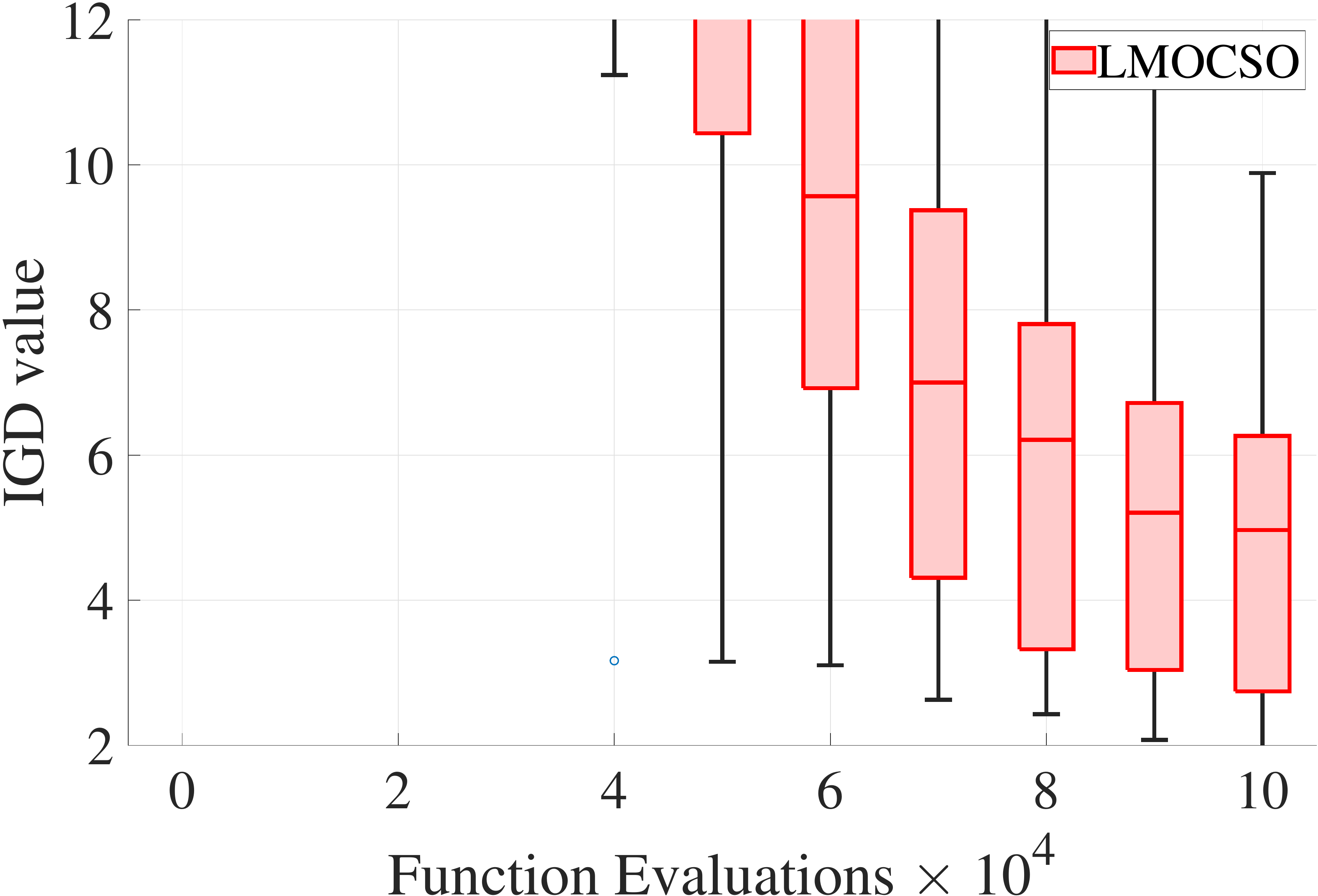}}
    \caption{Convergence profiles of the IGD values obtained by two representative large-scale MOEAs on tri-objective LSMOP3 and LSMOP6 with 100 decision variables and different initial population.}
    \label{fig:pre1}
\end{figure*}

\begin{figure*}[htbp]
    \centering
    \subfloat[{CCGDE3 Solves LSMOP1}]{\includegraphics[width=0.24\hsize]{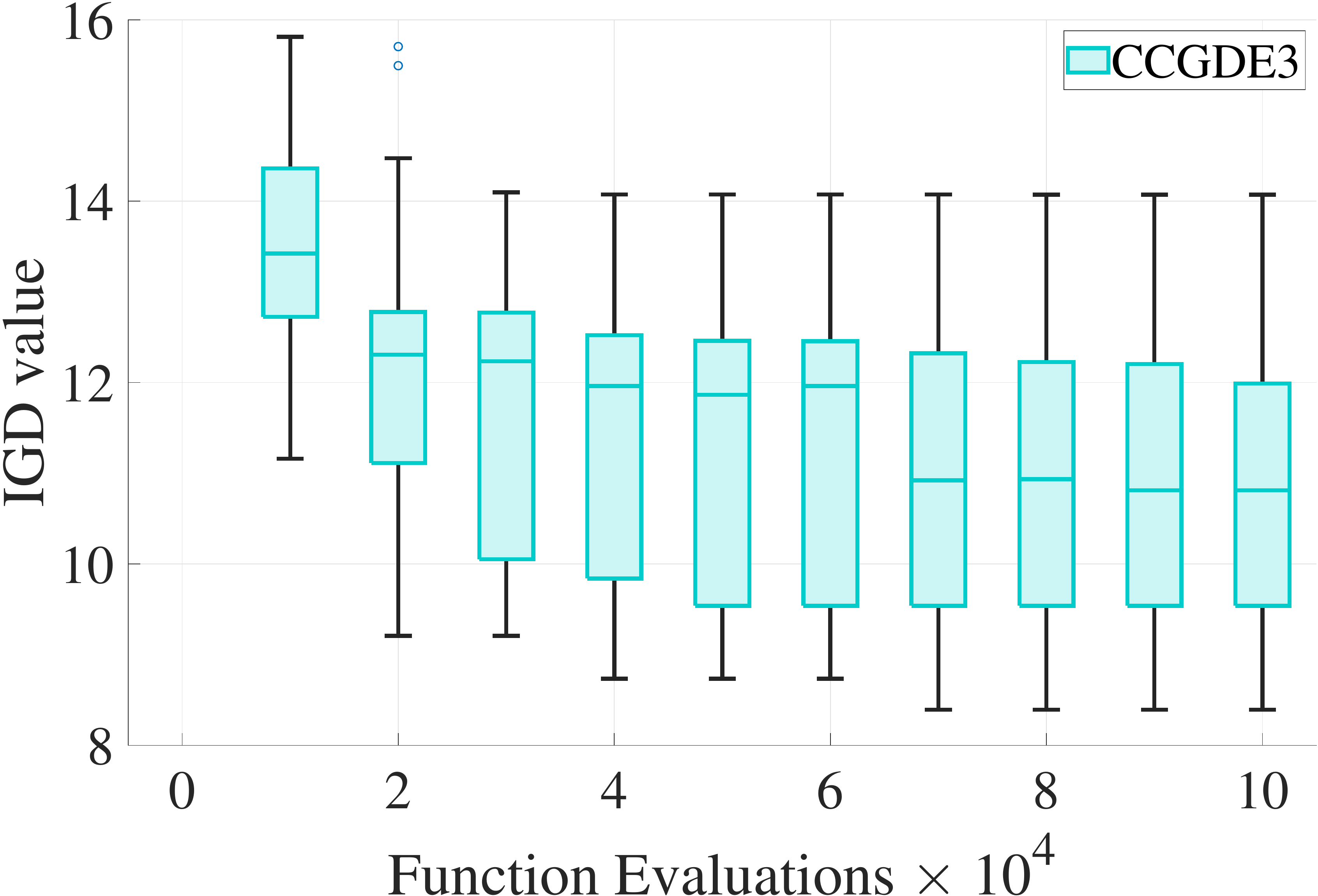}}
    \subfloat[{CCGDE3 Solves LSMOP6}]{\includegraphics[width=0.24\hsize]{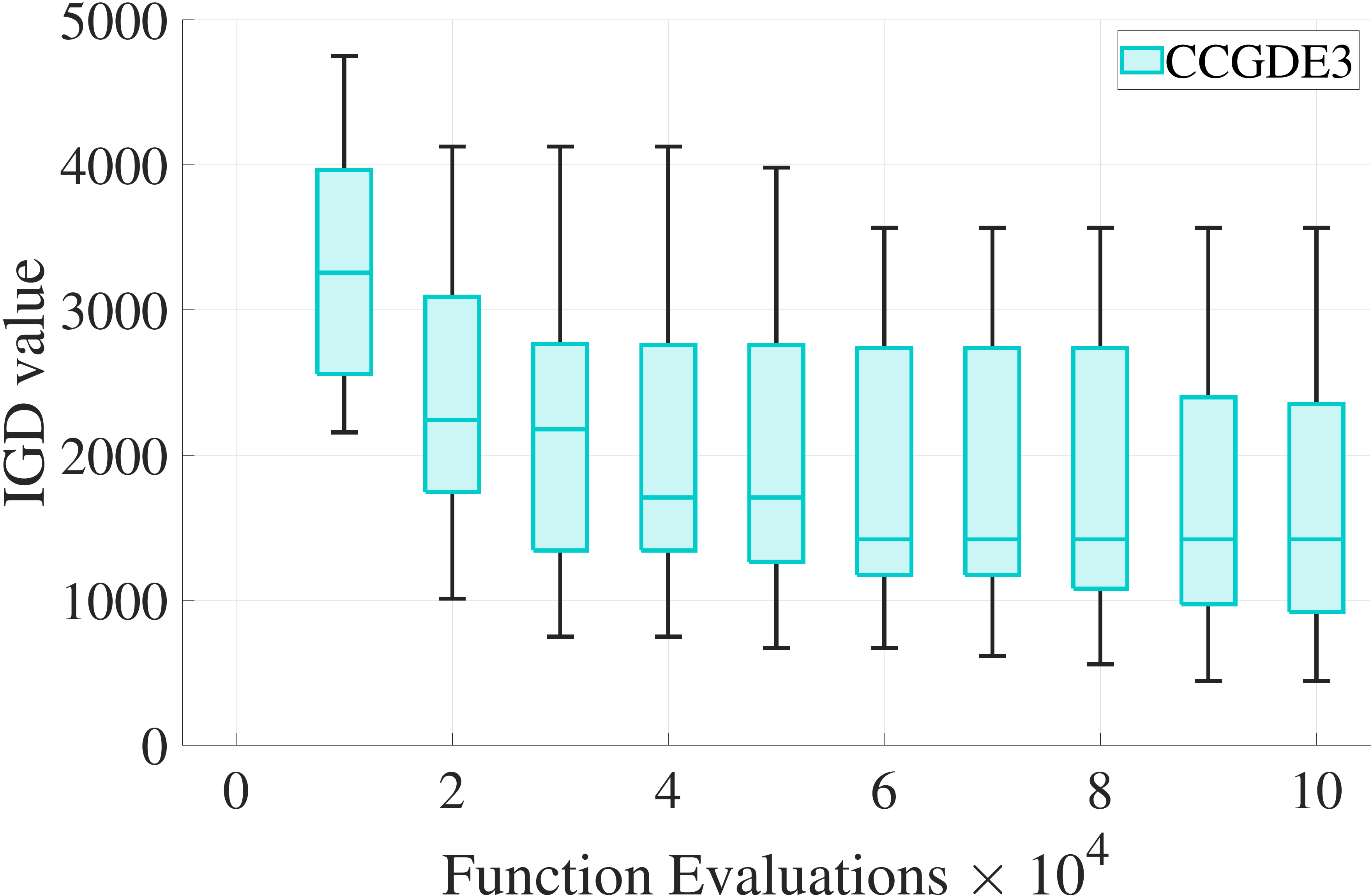}}
    \subfloat[{LMOCSO Solves LSMOP3}]{\includegraphics[width=0.24\hsize]{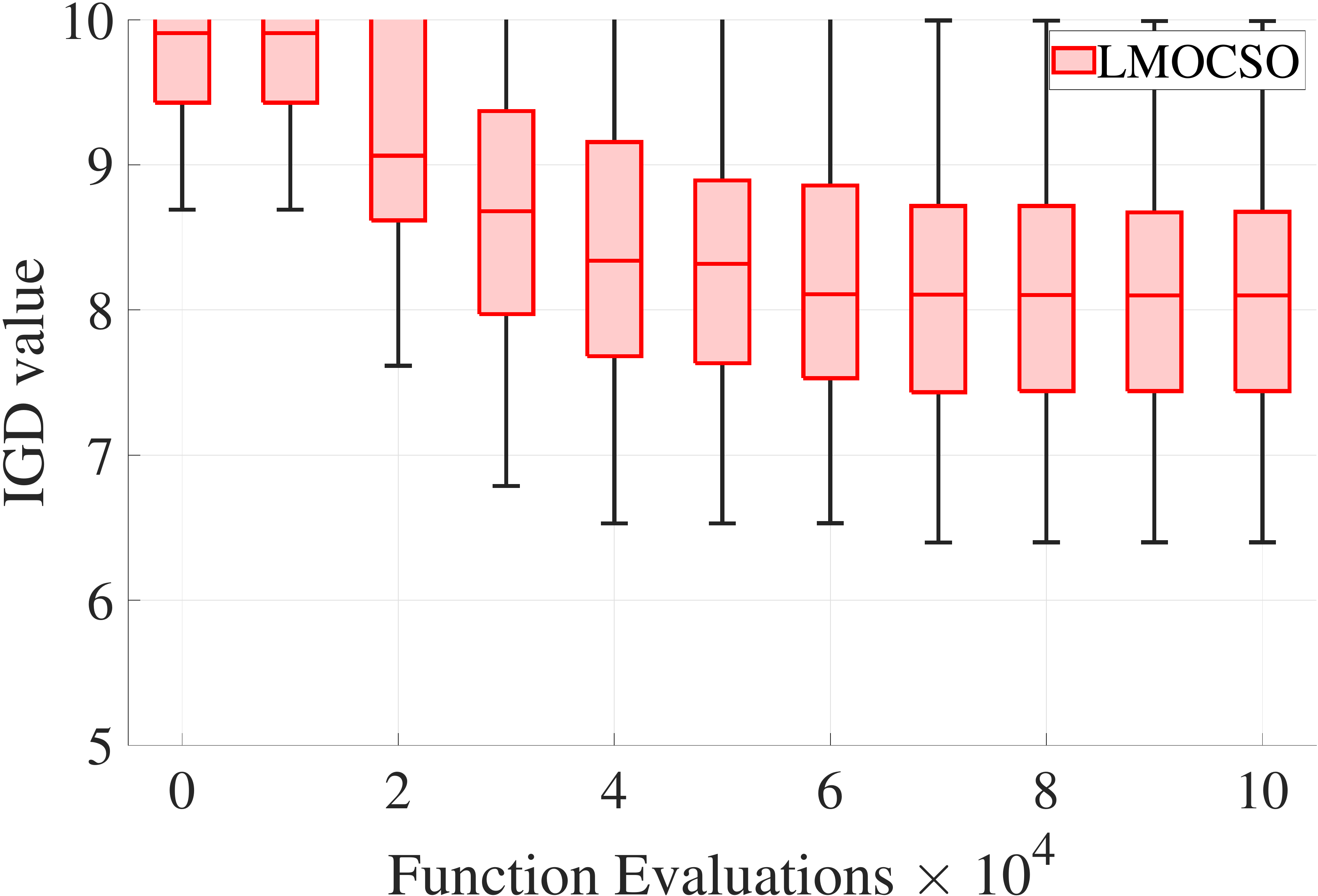}}
    \subfloat[{LMOCSO Solves LSMOP6}]{\includegraphics[width=0.24\hsize]{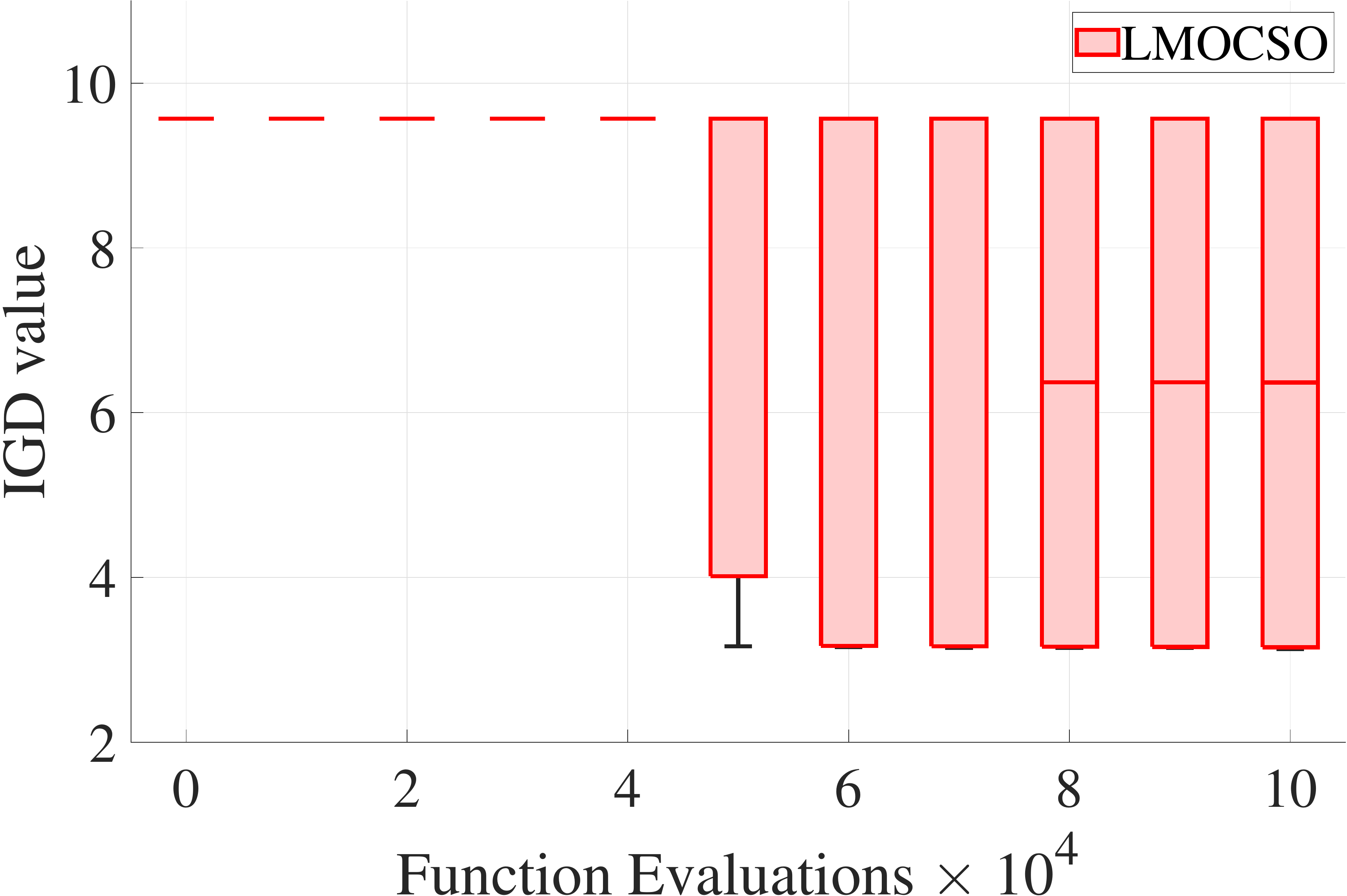}}
    \caption{Convergence profiles of the IGD values obtained by two representative large-scale MOEAs on tri-objective LSMOP3 and LSMOP6 with 100 decision variables and same population.}
    \label{fig:pre2}
\end{figure*}

{If we consider LSMOPs from the perspective of industrial applications, it is not difficult to find that a crucial point is obtaining acceptable and insensitive solutions within reasonable computational resources \cite{doi:10.1080/10286608.2012.672412, 4594481}. The point leads to a critical issue that is unsettled: the existing evolutionary algorithms are not only sensitive to different hyperparameters, such as the size of the population and the choice of the initial population, but the inherent randomness of EAs also makes the results of multiple runs tend to be quite different, even in the case that all hyperparameters are identically set. This problem leads to the common use of multiple runs strategies to obtain more insensitive solutions. However, a well-known disadvantage of evolutionary algorithms is the slow running speed; thus, the strategy of multiple runs further deteriorates this problem, which has inadvertently limited the applications of evolutionary optimization in some fields that need to obtain acceptable solutions quickly. Therefore, this leads to an urgent necessity to develop a performance-insensitive algorithm; that is, evolutionary optimization algorithms need only one run to obtain reliable and insensitive solutions. Research on improving the insensitivity of EAs will greatly expand the scope of such algorithms to solve real-world problems.}
\par
{Solving LSMOPs needs to deal with a larger search space, which further emphasizes the requirement of insensitivity to the algorithm’s performance. The algorithm needs to be able to obtain a reliable solution after only a few independent runs or even only one run of calculation. Unfortunately, the existing algorithms for LSMOPs still have room for improvement in performance insensitivity, which directly limits the application scope of these algorithms.} Inspired by the work in \cite{ARORA2012491} by Arora, we define the performance insensitivity of an algorithm as being able to converge from an arbitrary initialization. Performance insensitivity is a critical aspect if we select an algorithm for practical applications \cite{ARORA2012575}; for that, the algorithm with good insensitivity saves the user's time in the long run and removes uncertainty about the optimum solution found.
\par
A preliminary set of experiments on the sensitivity of the optimization results is shown in Figs. \ref{fig:pre1} and \ref{fig:pre2}. {We run the representative algorithms mentioned above (i.e., CCGDE3 and LMOCSO) 20 times and use IGD (inverted generational distance) \cite{RN98} to depict the performance.} The positions of the two ends of the rectangle in the figure correspond to the upper and lower quartiles of the IGD value obtained by the algorithm across 20 runs, and the horizontal line in the rectangle is the median line. {For each algorithm, we use 20 different populations (Fig. \ref{fig:pre1}) and the same population for 20 runs (Fig. \ref{fig:pre2}) as initialization.}
\par
It can be seen from Figs. \ref{fig:pre1} and \ref{fig:pre2} that whether it is the same initial population or different, two state-of-the-art large-scale MOEAs show inconsistent results over multiple runs. The sensitivity revealed by the experiments shows that the current algorithms are difficult to apply in many real-world fields. {For problems such as the capacitated arc routing problem (CARP) \cite{7533424} and intelligent transportation systems \cite{8315121}, the optimization algorithm should ensure that the deployed algorithm can produce stable and insensitive optimization results over multiple runs.}
\par
{Preliminary experiments not only show the necessity of improving the algorithms' performance insensitivity but also reveal some possible reasons for this phenomenon. Fig. \ref{fig:pre1} shows that, due to the randomness of initialization, an underperforming population can cause the population to get stuck in local optima, resulting in fluctuation and sensitivity. In addition, Fig. \ref{fig:pre2} shows that even with the same initial population, due to the stochastic nature of MOEAs, the quality of optimization results varies and leads to sensitive optimization results.}
\par
In this paper, we propose a large-scale multiobjective optimization algorithm for solving LSMOPs, which focuses on improving the performance insensitivity via Monte Carlo tree search (MCTS), in short, for LMOMCTS. In the proposed method, a specific instantiation of MCTS, the upper confidence bound applied to tree (UCT) \cite{RN220}, is used to generate a search tree with the population as nodes. {The parent node will randomly sample some decision variables for optimization and create a population for child nodes. Each child node will be evaluated to select the most suitable node for evolution.} Specifically, the main contributions of this paper are summarized as follows.
\begin{enumerate}
    \item {The major contribution is the proposed LMOMCTS for improving the performance insensitivity when solving LSMOPs. The algorithm is based on the tree search design to select nodes with high evaluation values for optimization, which can effectively avoid underperformed populations and thereby improve the algorithm's performance insensitivity while ensuring the performance for solving real-world applications.} 
    \item Aiming at the curse of dimensionality caused by large-scale decision variables, the proposed design samples different decision variables on a parent node throughout the evolutionary search so that the algorithm can search on different decision variables and alleviate the influence of the curse of dimensionality.
    \item {We systematically analyze and define a key metric for optimization algorithms, named performance insensitivity. We design two metrics to characterize performance insensitivity, providing some groundwork for follow-up research on this topic.}
\end{enumerate}

\par
The rest of the paper is organized as follows. In Section \ref{sec:Preliminaries-and-Related}, we will first introduce the preliminaries and related works about solving LSMOPs and briefly review the MCTS, an essential tool to address the performance insensitivity. Our proposed LMOMCTS method is presented in Section \ref{sec:LMOMCTS} in detail. In Section \ref{sec:exp}, we conduct a series of experiments between LMOMCTS and chosen state-of-the-art large-scale MOEAs on LSMOPs. The last section highlights some conclusive discussions and draws future research directions.

\section{Preliminary Studies and Related Work}
\label{sec:Preliminaries-and-Related}
\subsection{Large-scale Multiobjective Optimization}

Large-scale multiobjective optimization problems can be mathematically formulated as follows:
\begin{equation}
\begin{aligned}
\textbf{minimize}\ \boldsymbol F(\boldsymbol x) =& (f_1(\boldsymbol x), f_2(\boldsymbol x),\dots,f_m(\boldsymbol x))\\
&s.t.\ \boldsymbol x \in \Omega
\end{aligned}
\end{equation}
where $\boldsymbol x=(x_1,x_2,...,x_d)$ is $d$-dimensional decision vector, $\boldsymbol F=(f_1,f_2,\dots,f_m)$ is $m$-dimensional objective vector. It is worth noting that large-scale optimization problems consider the dimension in $d$ to be greater than $100$ \cite{8681243}.
\par
Suppose $\boldsymbol x_1$ and $\boldsymbol x_2$ are two solutions of an MOP, solution $\boldsymbol x_1$ is known to Pareto dominate solution $\boldsymbol x_2$ (denoted as $\boldsymbol x_1 \prec \boldsymbol x_2$), if and only if $f_i(\boldsymbol x_1) \leqslant f_i(\boldsymbol x_2) (\forall i = 1,\dots,m)$ and there exists at least one objective $f_j (j \in \{1, 2, \dots , m\})$ satisfying $f_j(\boldsymbol x_1) < f_j(\boldsymbol x_2)$. The collection of all the Pareto optimal solutions in the decision space is called the Pareto optimal set (PS), and the projection of PS in the objective space is called the Pareto optimal front (PF).

\subsection{The Monte Carlo Tree Search Method}
The Monte Carlo method \cite{RN57}, also called Monte Carlo sampling, originated from statistical physics and is a kind of stochastic algorithm. The idea of the Monte Carlo method is to use repeated sampling to obtain the distribution feature of a random variable and use it as the solution to the problem and has been applied to solve MOPs {in many fields}\cite{HAZRA2019454, BARNOON20222747}.
\par
Monte Carlo tree search (MCTS) \cite{6145622} is a category of algorithms that combines the Monte Carlo search with game tree search. The upper confidence bound applied to tree (UCT) is a classical instantiation of MCTS \cite{RN220}, which applies bandit ideas to guide Monte Carlo planning.
\par
\par
\begin{figure}[ht]
    \centering
    \includegraphics[width=1\hsize]{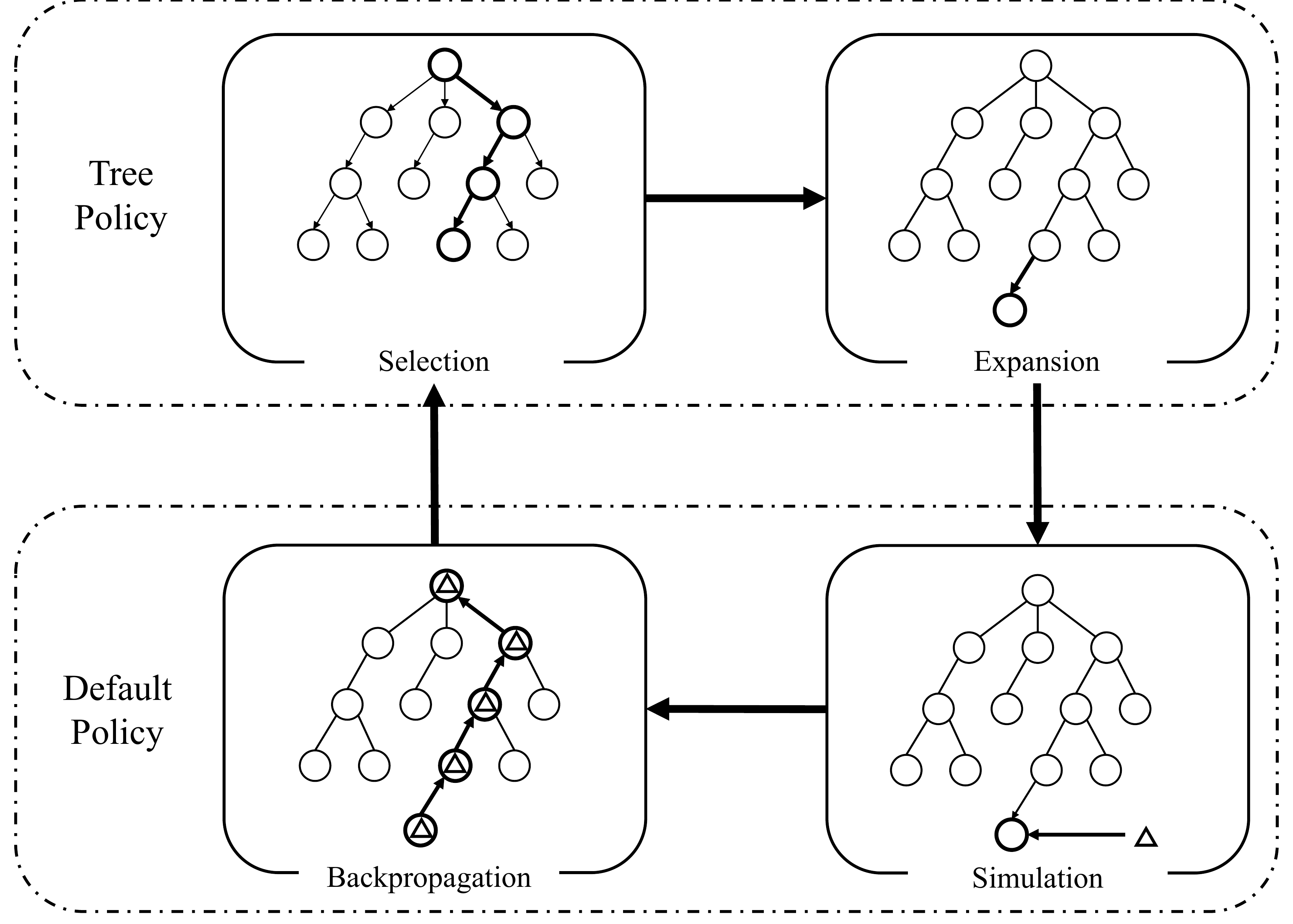}
    \caption{The basic process of UCT. Selection: Select the node with the highest UCB value and is not fully expanded; Expansion: Expand the node; Simulation: Use Default Policy to evaluate the node; Backpropagation: Backward propagation of evaluation to all parent nodes to update UCB values.}
    \label{fig:UCT}
\end{figure}
The UCT is an extended realization of {the upper confidence bound (UCB)} to the search tree. The general idea of UCT is to treat each node in the search tree as an agent, and its child nodes correspond to the gambling machines. {When selecting nodes, we choose based on the UCB value. Fig. \ref{fig:UCT} shows the iterative process of UCT iteration}: tree policy and default policy. In the tree policy stage, starting from the root node, the child node with the largest UCB value is selected unless the node is entirely expanded; then, a new node is expanded from this node. {In the default policy phase, a simulation is performed on the new node as the evaluation value; then, the value is propagated back to update the UCB value of each node on the path.} Iterate repeatedly until the budget of computing resources is reached, and finally, return the child node with the highest value under the root node.
\par
Based on the preliminary experiments in the introduction, the proposed algorithm generates multiple nodes in the early stage of evolution. Different nodes represent different evolution directions of the population. Then, the algorithm evaluates each node, and nodes with low evaluated values will be selected and evolve less. The design avoids local optima in some directions and improves the performance insensitivity of the algorithm. In the proposed framework, there are two methods involving Monte Carlo. The first is to use Monte Carlo tree search to search for potential populations for evolution. The second is to use the Monte Carlo method \cite{fasthv} to estimate the hypervolume\cite{10.5555/1762545.1762618} of the population during the evaluation of the Monte Carlo tree search.
\begin{figure*}[ht]
    \centering
    \includegraphics[width=0.95\hsize]{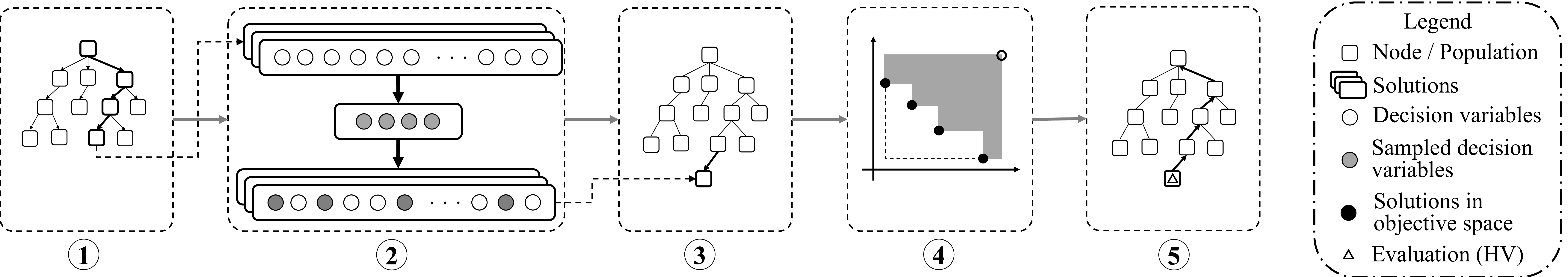}
    \caption{\textbf{Overview of the LMOMCTS framework.} The proposed method constructed a tree in which each node is a population. 1. The UCT guides the method to select a promising node. 2. The algorithm samples decision variables for evolution. 3. The algorithm constructs a child node based on the evolved population. 4. The algorithm estimates Hypervolume as the evaluation of the node. 5. The algorithm updates the UCT via the estimated evaluation.}
    \label{fig:Framework}
\end{figure*}
\subsection{Related Works}
Large-scale MOEAs can be categorized into three different groups: decision variable grouping-based, decision space reduction-based, and novel search strategy-based methods \cite{tian2021evolutionary}.
\par
\subsubsection{Decision Variable Grouping}
The first MOEA for solving LSMOPs is CCGDE3 proposed by Antonio \emph{et al.} \cite{6557903}, which maintains several independent subgroups. {Another technique in this category is based on variable analysis.} Ma \textit{et al.} \cite{7155533} proposed a multiobjective evolutionary algorithm based on decision variable analysis (MOEA/DVA). Zhang \emph{et al.} \cite{RN82} proposed a large-scale evolutionary algorithm (LMEA) based on the clustering of decision variables.
\par
In \cite{RN259}, an adaptive dropout on decision variables was proposed, which took advantage of the significant differences between different solution sets in the decision space to guide the selection of some crucial decision variables. Different from this algorithm, our method uses random sampling decision variables for optimization, which cooperate with UCT to ensure the performance of the algorithm.
\par
The disadvantage of decision variable grouping-based methods is that they may incorrectly identify the linkages between decision variables \cite{RN82}, which may lead to local optima and high sensitivity with multiple runs.
\par
\subsubsection{Decision Space Reduction} In this category, problem transformation-based MOEAs are suggested to solve LSMOPs.{ Zille \emph{et al.} \cite{RN89} proposed a weighted optimization framework (WOF) for solving LSMOPs. In \cite{CAO2020100626}, graph-based differential grouping was suggested to decompose the variables in LSMOPs.}
\par
Decision space reduction-based methods may lose important information about the original search space after reduction. Therefore, the original large-scale search space cannot be fully explored \cite{9723458} and may produce sensitive results.
\par
\subsubsection{Novel Search Strategy}
The third category is based on enhanced search. Tian \emph{et al.} \cite{8681243} used a competitive group optimizer to solve LSMOPs. {Recently, novel search strategies, such as generative adversarial networks\cite{RN90}, self-exploratory competitive swarm optimization\cite{QI20221601}, adaptive competitive swarm optimization with inverse modeling \cite{GE20221441}, and probabilistic prediction models \cite{RN257} were introduced to solve LSMOPs.}
\par
{Our algorithm can be regarded as a method to determine the best search directions. However, compared to using the probabilistic prediction model, GAN, and CSO to determine the direction, our proposed method has the advantage of maintaining the tree, by which multiple directions can be selected in a certain generation.}
\par
Based on the above discussions, it can be observed that many large-scale MOEAs have been proposed to deal with LSMOPs. However, the abovementioned approaches ignore the requirement of solving the LSMOP in industrial practice, which is to obtain insensitive optimization results. In response to this need, we present an algorithm for improving the performance insensitivity when solving the LSMOPs.
\section{Large-scale Multiobjective Optimization algorithm via Monte Carlo Tree Search (LMOMCTS)}
\label{sec:LMOMCTS}
\subsection{Overview}
The basic idea of the proposed algorithm regards the evolution process as a game search process, using UCT to generate a tree to search for the best offspring population. Each node on the search tree represents an instant population. Due to the large-scale decision variables of the LSMOPs, a method that cooperates with MCTS, decision variable sampling for optimization, is applied to sample decision variables to be searched from the parent node to form a new node. Fig. \ref{fig:Framework} illustrates the proposed framework.
\par
\subsection{Decision Variable Sampling for Optimization}
Inspired by decision variable grouping and reduction designs, we believe the decision variable is the key to addressing the LSMOP. Furthermore, given that our framework is based on the MCTS, random sampling is preferred. Therefore, the proposed algorithm samples different decision variables on the parent node for optimization to obtain different child nodes. The designed method of sampling different decision variables to evolve reduces the impact of the large-scale search space and ensures the algorithm's performance.
\par
\begin{figure}[ht]
    \centering
    \includegraphics[width=0.95\hsize]{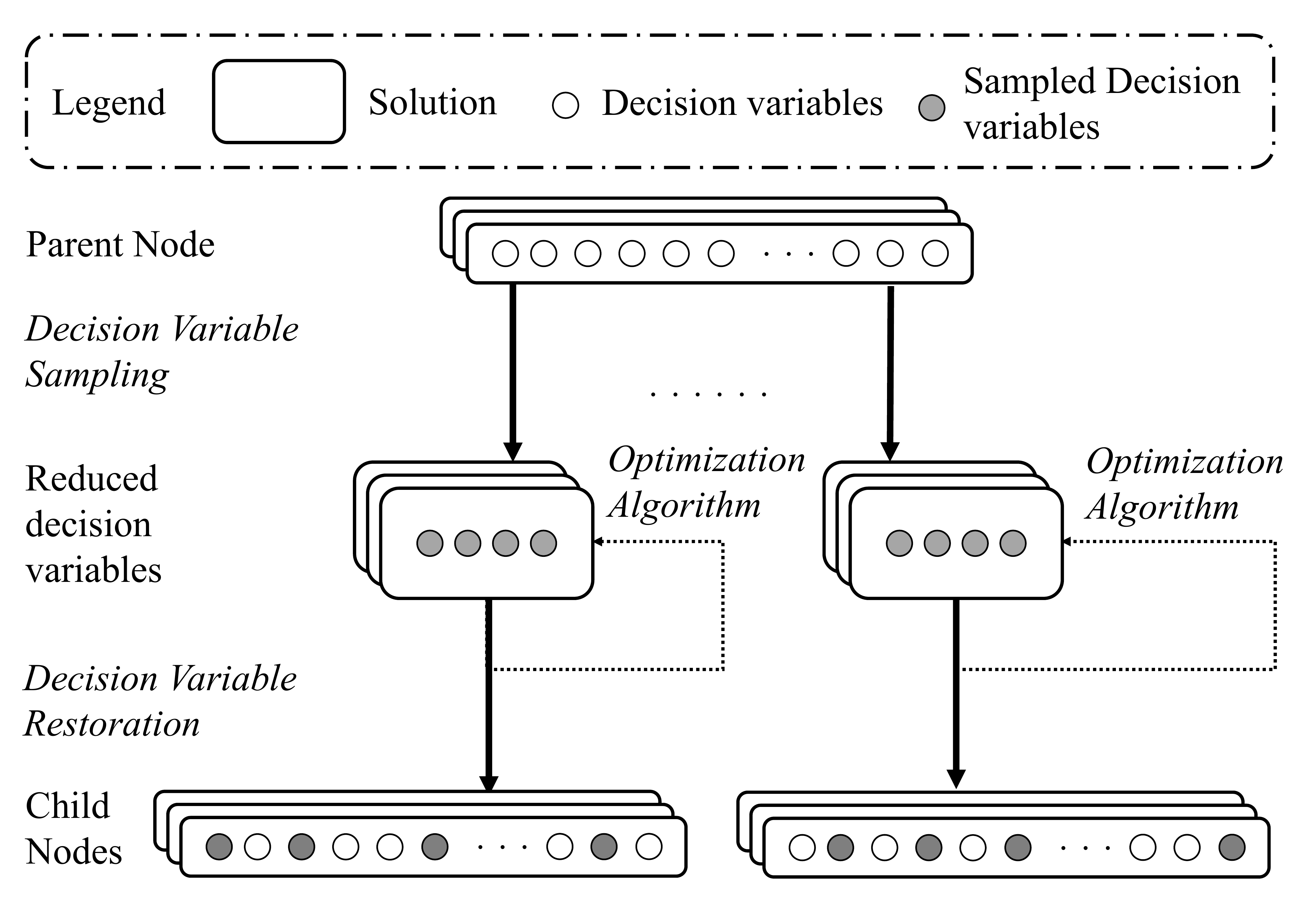}
    \caption{\textbf{Decision variable sampling for optimization.} Each rounded box is a solution, and the circle within corresponds to the decision variable. A subset of decision variables is first randomly sampled, and each solution is dimensionally reduced. An evolutionary algorithm is then used for the optimization on these decision variables. Finally, according to the sampled decision variables, all solutions are restored to the whole population and constitute the child node.}
    \label{fig:Expanding}
\end{figure}
Given a parent node $N_p$ and its corresponding population $P$ with $n$ solutions, and each solution has $d$ decision variables, for each child node of $N_p$, the algorithm samples $d_n$ decision variables from $d$ decision variables for optimization. Other decision variables that are not sampled remain the same for generating this child node. They are only used for evaluation.
\par
Based on sampling decision variables, each child node of the parent node $N_p$ will optimize different decision variables based on the parent population. After obtaining the sampled $d_n$ decision variables, we use traditional optimization algorithms to perform the evolutionary search on these decision variables. Each solution of the parent population is first reduced to the $d_n$ dimension according to the sampled decision variables. Afterward, the algorithm performs an evolutionary strategy on the reduced population according to the optimization algorithm used and is then restored to a $d$-dimensional solution. The offspring are evaluated, and environmental selection is performed. After several iterations, the final population is used to construct child nodes $N_c$. Fig. \ref{fig:Expanding} illustrates the method of decision variable sampling for optimization. The algorithm of decision variable sampling and optimization is presented in Algorithm \ref{alg:dvso}. In the process of performing the optimization algorithm on $P$ (Line 8), reproduction operators are performed on the reduced solution, and the reduced solution will be restored to complete solution for evaluation and selection.
\par 
\begin{algorithm}
    \caption{Decision Variable Sampling for Optimization (DVSO)}
    \label{alg:dvso}
    \KwIn{
        Parent Node $N_p$; Optimization algorithm $A$; Size of decision variable sampled $d_n$; Function evaluation $e$.
    }
    \KwOut{Child Node $N_c$}
    $P \gets$ Population of node $N_p$ \;
    $d \gets$ Size of decision variable \;
    Randomly sample $d_n$ decision variables from $d$ decision variables \;
    \For{$ \boldsymbol x \in P$}{
        $\boldsymbol x' \gets $ Reduced to $d_n$-dimension vector \;
        {$P' \gets x'$} \;
    }
    {$P' \gets A(P', e)$} // \textit{Optimize $P'$ with  $A$ for $e$ evaluations, using P' as a initial population} \;
    Restore $\boldsymbol x'$ for each $\boldsymbol x'\in P'$ and put into {$P_c$}\;
    $N_c\gets$ Construct child node based on {$P_c$} \;
    \Return Child node $N_c$ \;
\end{algorithm}
The use of the UCT ensures that the algorithm will select a more educated population in each subsequent iteration to ensure convergence. {On this basis, the use of the random sampling method accelerates the speed of generating child nodes. Furthermore, the sampling on decision variables will not be limited to certain decision variables, which provides multiple sampling on a node. For each node, different child nodes with different sampled decision variables are generated, which in turn avoids bad grouping and dimension reduction on a single child node and increases the diversity of the population.} Sampling different decision variables in different child nodes to search and restore avoids the loss of information in the decision space that may be caused by dimension reduction.
\par
`
\subsection{The proposed LMOMCTS}
The process of expanding the node is decision variable sampling for optimization. Another important part of the framework is the simulation, which evaluates the suitability of the node for further searches.
\par
{The evaluation of a node can be measured by the population's hypervolume (HV) value. HV is a promising performance indicator \cite{10.5555/1762545.1762618} to guide the search process. This indicator gives the hypervolume of the dominated portion of the objective space and is strictly monotonic\cite{1197687}. In addition, the hypervolume measure guarantees that any approximation set PS that achieves the maximally possible quality value for a particular problem contains all Pareto-optimal objective vectors\cite{10.1007/3-540-36970-8_37}, which ensures the diversity of the population. These properties make it well suited to be the evaluation policy in the MCTS framework.}
\par
However, in the evaluation policy of our framework, the exact hypervolume indicator values are not crucial. What matters is the corresponding ranking, i.e., whether the evaluation of one node (population) is larger than that of another node (population). To this end, a faster hypervolume approximation method via Monte Carlo sampling proposed in \cite{fasthv} is adopted.
\par
{LMOMCTS uses the UCB value to decide the node for expansion. The UCB value consists of the estimated HV value (exploitation) and the value of the exploration. Given a parent node $N$ and a child node $N'$ that has been expanded but has not calculated the UCB value, the UCB of $N'$ is calculated as:}
\begin{equation}
\label{equ:UCB}
\begin{aligned}
{\text{UCB}(N') = \Delta(N') + \sqrt{ ({2\ln{t}}) / {t(N')}}}
\end{aligned}
\end{equation}
{where $t(N')$ is the number of times node $N'$ has been selected, and $t$ is the sum of the number of times the children of node $N$ have been selected. UCB is a trade-off between the estimated HV value and the number of times the child node $N'$ is selected, and its role is to balance the exploitation and exploration on each node.}
\par
After the above introduction, we then present the main steps of LMOMCTS. The specific implementation of the LMOMCTS algorithm is given in Algorithm \ref{alg:LMOMCTS}. First, we initialize the population as the root node, and the evaluation $\Delta$ of the root node is set to $0$. {Then, the algorithm starts from the root node and selects the node with the largest UCB value each time. The selection will stop at the node with fewer than $k$ child nodes (Lines 8-16).} This step is to select promising nodes that can still expand child nodes. $N \gets \arg \max_{N' \in \text{children\ of}\ N} \text{UCB}(N')$ selects the node with the highest UCB value from all child nodes of $N$.
\par
Then, we use decision variable sampling for optimization to construct a new node $N_c$ (Line 17). Afterward, we estimate the hypervolume value of the child node $N_c$ as evaluation $\Delta$, then backpropagate $\Delta$ and update UCB values (Lines 18). Decision variable sampling for optimization is the algorithm presented in Algorithm \ref{alg:dvso}, and Monte Carlo-based HV is proposed in \cite{fasthv}. The algorithm updates the archived node with $N_c$ if the hypervolume of the child node $N_c$ is higher than that of the archived node (Lines 19 to 22).  {Line 23 discards the population of the node that has $k$ child nodes and is not $N_A$, which saves memory since such a node no longer needs to be evolved or evaluated.} Lines 24 to 28 update the UCB values of all parent nodes of child node $N$. The algorithm will repeat until the computing resource is exhausted.
\par
\begin{algorithm}
    \caption{LMOMCTS}
    \label{alg:LMOMCTS}
    \KwIn{
        $f$ (The ratio of the sampling decision variables); Optimization algorithm $A$; $E$ (Function evaluation); $e$ (Function evaluation for \textbf{DVSO}).
    }
    \KwOut{$A$(Archived node with best HV)}
    $P \gets$ Initial population \;
    $d \gets$ Size of decision variables \;
    $d_n \gets d \cdot f$ \; 
    $k = \lceil - 1 / (d_n \log_{10}(1- 1 / d))  \rceil$ \;
    Create root node $N_0$ based on $P$ \;
    $\text{UCB}(N_0) \gets 0$; $N_A \gets N_0$ \;
    \While {within function evaluation $E$} {
        $N \gets N_0$ \;
        \While {The number of children of $N$ reaches $k$} {
            $t \gets \sum_{N' \in \text{children\ of}\ N} t(N')$ \;
            \For {Child $N'$ of $N$} {
                $\text{UCB}(N') \gets \Delta(N') + \sqrt{ ({2\ln{t}}) / {t(N')}}$ \;
            }
            $N_p \gets \arg \max_{N' \in \text{children\ of}\ N} \text{UCB}(N')$\;
            $t(N) \gets t(N) + 1$ \;
        }
        $N_c \gets$ \textbf{DVSO} $(N_p, A, d_n, e)$ \;
        $\Delta(N_c) \gets $ \textbf{HV}($N_c$) \cite{fasthv} \;
        \If {$\Delta(N_A) < \Delta(N_c)$} {
            // \textit{Update archived node}\;
            $N_A \gets N_c$ \;
        }
        Discard the population of the node that has $k$ child nodes and is not $N_A$ \;
        \While {$N_p \neq N_0$} { // \textit{Backpropagate $\Delta(N_c)$} \;
            $\Delta(N_p) \gets \Delta(N_p) + \Delta(N_c)$ \;
            $N_p \gets \text{Parent node of\ } N_p$ \;
        }
    }
    \Return $A$ \;
\end{algorithm}

In the specific implementation, to make each decision variable have at least a 90\% chance of being sampled if sampling from $d$ decision variables where each time $d_n$ decision variables are sampled, the branching factor (the number of child nodes of a parent node) $k$ should satisfy:

\begin{equation}
\begin{aligned}
\left ( 1- 1/d \right )^{d_n k} & \leq 1 -0.9 \\
k &  \geq -  1 / (d_n \log_{10} (1- 1/ d))
\end{aligned}
\end{equation}

Therefore, branching factor is calculated as follows:

\begin{equation}
\label{equ:factork}
\begin{aligned}
k = \lceil - 1 / (d_n \log_{10}(1- 1 / d))  \rceil
\end{aligned}
\end{equation}
where $\lceil x \rceil$ is the ceiling function that maps $x$ to the least integer greater than or equal to $x$.
\par
The parameter $f(f=d_n/d)$ is the sampling ratio. Using the ratio instead of a fixed value prevents the number of child nodes from being too large due to the small number of samples. $N_A$ is the archived node with the best HV.
\subsection{{Complexity Analysis}}
{The complexity of the Decision Variable Sample for Optimization module of the LMOMCTS algorithm is $\mathcal{O}(d\cdot n + d_n\cdot n\cdot m)$, where $d$ is the number of decision variables, $d_n$ is the number of sampled decision variables, $n$ is the population size and $m$ is the number of objectives.}
\par
{The runtime of the the proposed algorithm can be computed as $\mathcal O(\log(I)\cdot k + I\cdot \mathcal O(D) + I\cdot \mathcal O(V))$, where $I$ is the number of iterations, which is $E/e$, k is the branching factor, $\mathcal O (D)$ is the complexity of explanding the node, which is $\mathcal{O}(d_n\cdot n\cdot m)$, and $\mathcal O (V)$ is the complexity of estimating the hypervolume, which $\mathcal O(n\cdot m)$. Therefore the complexity of LMOMCTS is $\mathcal O(\log(I)k + I\cdot d_n\cdot n\cdot m + I\cdot n\cdot m) = \mathcal O(Id_nnm)$}
\par
{In each iteration, we generate one node with $n$ $d$-dimensional solutions, and we discard nodes that have $k$ child nodes. Thus the memory complexity is $\mathcal O(\log (I)\cdot n \cdot d)$.}
\section{Experiments}
\label{sec:exp}
\subsection{Algorithms in Comparison and Test Problems}
{The proposed LMOMCTS is compared with several state-of-the-art large-scale MOEAs, including NSGA-II\cite{996017}, CCGDE3 \cite{6557903}, MOEA/DVA\cite{7155533}, LMEA \cite{RN82}, WOF \cite{RN89},  S3-CMA-ES \cite{CHEN2020457}, LMOCSO \cite{8681243} and GMOEA \cite{RN90}.}
\par
The experiments are conducted on the widely used test problems, LSMOP1 - LSMOP9 \cite{7553457}. Specifically, LSMOP1 – LSMOP4 have a linear PF, LSMOP5 – LSMOP8 possess a convex PF, and LSMOP9 has a disconnected PF.
\par
For each test problem, the number of objectives is three, and the decision variable numbers vary from 100 to 5,000 ($d \in \{100,200,500,1,000,2,000,5,000\}$). For each test instance, each algorithm is run independently 20 times, and Wilcoxon rank-sum \cite{Haynes2013} is used to compare the statistical results obtained by LMOMCTS and the compared algorithms at a significance level of 0.05. In the tables, +/=/- indicate that compared algorithms perform significantly better, indifferent, or significantly worse than the proposed LMOMCTS, respectively, from a statistically meaningful sense.

\subsection{Performance Indicators}
This study uses the following metrics to evaluate the performance of the compared algorithms.
\subsubsection{Inverted Generational Distance (IGD)} IGD \cite{RN98} is a metric for quantifying the convergence of the solutions. The smaller the IGD value is, the better the performance on the convergence of the solution. 
\subsubsection{Hypervolume (HV)} The hypervolume \cite{1583625} of a set of solutions measures the size of the portion of objective space that is collectively dominated by those solutions. .
\subsubsection{Performance Insensitivity} To measure the performance insensitivity of the algorithm in terms of convergence and diversity, we propose two indicators, insensitive-IGD and insensitive-HV, which are defined as follows:
\begin{equation}
\label{eq:insensitiivity}
    \begin{aligned}
    \text{insensitive}&\text{-IGD}(\mathcal A) = \\
    &\frac{\sum_{n_r}(\text{IGD}(\mathcal A) - \min_{\mathcal A' \in \mathbb A}(\text{IGD}(\mathcal A')))^2}{n_r} \\
    \text{insensitive}&\text{-HV}(\mathcal A) = \\
    &\frac{\sum_{n_r}(\text{HV}(\mathcal A) - \max_{\mathcal A' \in \mathbb A}(\text{HV}(\mathcal A')))^2}{n_r}
    \end{aligned}
\end{equation}
\par
These indicators are based on variance. Nevertheless, insensitive-IGD and insensitive-HV do not describe the dispersion of the optimization results of a single algorithm on a problem but the degree of dispersion of the algorithm from the best optimization result. We do not use variance directly because some algorithms may converge densely to poor results multiple times. Algorithms that perform poorly many times but have small variance are not of practical value.
\par
Smaller values of insensitive-IGD and insensitive-HV indicate that the algorithm is less sensitive. That is, the performance of multiple runs is good and consistent.
\par
In Eq. (\ref{eq:insensitiivity}), $\mathcal A$ is the measured algorithm. $n_r$ is running times, and in our experiment, $n_r$ is 20. $\text{IGD}(\mathcal A)$ is the IGD value obtained by algorithm $\mathcal A$, $\text{HV}(\mathcal A)$ is the HV value obtained by algorithm $\mathcal A$, $\mathbb A$ is the set of all compared algorithms.
\subsection{Parameter settings}
For a fair comparison, the recommended parameter settings adopted for the compared algorithms have achieved the best performance as reported in the original literature and are presented in the Table S-I of the supplementary materials. All algorithms are implemented in PlatEMO \cite{RN92}.
\subsubsection{Population Size} Considering the settings of the population size in \cite{8681243,RN90}, in our experiment, the population size is set to 300 for test instances with three objectives.
\subsubsection{Parameter Settings of LMOMCTS} The ratio of sampling decision variables $f$ is set to 0.2. The function evaluation $e$ for decision variable sampling is set to $0.01 \times E$. NSGA-II is adopted as the optimization algorithm in Algorithm \ref{alg:dvso} as $A$.
\subsubsection{Termination Condition} {The number of total function evaluations, $E$, is set to 100,000, which is practical for real-world applications \cite{RN88}.}
\begin{table*}[htbp]
\scriptsize
  \centering
  \caption{Insensitive-IGD Values Obtained By 9 Compared Algorithms on 54 Test Instances From LSMOP Test Suite With Tri-Objective. The Best Result in Each Row is Highlighted in Bold.}
    \begin{tabular}{ccccccccccc}
    \toprule
    Problem & D     & NSGA-II & CCGDE3 & MOEA/DVA & LMEA  & WOF   & S3-CMA-ES & LMOCSO & GMOEA & LMOMCTS \\
    \midrule
    \multirow{6}[1]{*}{LSMOP1} & 100   & 3.26e-02= & 1.17e+01- & 2.94e+00- & 1.91e+00- & 4.61e-02= & 8.86e-01- & 1.56e-01- & \textbf{9.49e-03=} & 2.96E-02 \\
          & 200   & 2.77e-01- & 3.37e+01- & 1.55e+00- & 9.44e+01- & 6.20e-02= & 3.90e+01- & 6.07e-01- & 6.33e-02= & \textbf{1.47e-02} \\
          & 500   & 3.23e+00- & 5.80e+01- & 7.22e-02= & 1.11e+02- & \textbf{2.11e-02=} & 3.72e+03- & 1.25e+00- & 7.97e+00- & 3.67E-02 \\
          & 1000  & 1.37e+01- & 7.36e+01- & \textbf{2.11e-06=} & 1.21e+02- & 5.88e-02= & 5.64e+03- & 1.61e+00- & 2.69e+03- & 4.60E-02 \\
          & 2000  & 5.20e+01- & 1.01e+02- & 2.26e-01- & 2.16e+00- & 1.64e-01- & 3.10e+02- & 1.86e+00- & 4.14e+03- & \textbf{9.04e-02} \\
          & 5000  & 8.01e+01- & 9.21e+01- & 7.49e+03- & 7.34e+03- & 9.69e-03= & 4.09e+02- & 1.26e+00- & 3.62e+03- & \textbf{1.14e-03} \\
    \midrule
    \multirow{6}[0]{*}{LSMOP2} & 100   & 9.30e-03= & 1.65e-02= & 8.42e-03= & 2.93e-04= & 5.93e-03= & 3.74e-04= & 2.28e-03= & 6.54e-03= & \textbf{1.96e-04} \\
          & 200   & 3.26e-03= & 4.13e-03= & 2.64e-03= & 3.76e-03= & 2.23e-03= & 5.79e-04= & 4.32e-04= & 2.89e-03= & \textbf{1.44e-05} \\
          & 500   & 4.53e-04= & 5.47e-04= & 1.83e-04= & 3.68e-04= & 1.45e-04= & 3.13e-01- & \textbf{1.38e-06=} & 7.77e+00- & 2.52E-06 \\
          & 1000  & 1.62e-04= & 1.88e-04= & 4.51e-05= & 1.49e-04= & 2.04e-05= & 5.21e+00- & \textbf{6.21e-08=} & 2.78e+03- & 2.57E-05 \\
          & 2000  & 1.16e-04= & 8.96e-05= & 2.45e+00- & 2.42e+00- & 1.41e-06= & 2.84e+00- & \textbf{6.91e-08=} & 4.18e+03- & 4.77E-05 \\
          & 5000  & 1.30e-04= & 8.59e-05= & 7.46e+03- & 7.44e+03- & 2.24e-08= & 5.70e+00- & \textbf{2.02e-09=} & 3.75e+03- & 5.85E-05 \\
    \midrule
    \multirow{6}[0]{*}{LSMOP3} & 100   & 1.76e-01- & 1.13e+02- & 4.83e+02- & 1.90e+01- & \textbf{4.14e-02=} & 2.75e+00- & 6.14e+01- & 1.34e-01- & 4.50E-02 \\
          & 200   & 4.75e+00- & 1.98e+02- & 3.76e+02- & 7.71e+04- & \textbf{1.52e-09=} & 1.22e+02- & 7.37e+01- & 5.44e-01- & 2.47E-09 \\
          & 500   & 6.04e+01- & 2.55e+02- & 1.65e+01- & 2.26e+05- & \textbf{2.51e-03=} & 5.03e+09- & 1.07e+02- & 3.49e+00- & 3.13E-03 \\
          & 1000  & 2.18e+02- & 3.41e+02- & 2.44e+00- & 1.35e+04- & \textbf{7.04e-03=} & 5.47e+04- & 1.48e+02- & 2.67e+03- & 8.74E-03 \\
          & 2000  & 3.26e+02- & 3.47e+02- & 5.09e+00- & 2.17e+00- & 3.82e-02= & 2.56e+07- & 1.28e+02- & 4.16e+03- & \textbf{3.82e-02} \\
          & 5000  & 5.98e+02- & 3.64e+02- & 7.27e+03- & 7.28e+03- & \textbf{0.00e+00=} & 1.29e+08- & 1.58e+02- & 3.51e+03- & \textbf{0.00e+00} \\
    \midrule
    \multirow{6}[0]{*}{LSMOP4} & 100   & 4.28e-02= & 9.41e-02- & 4.55e-02= & \textbf{6.20e-04=} & 3.20e-03= & 3.55e-02= & 2.53e-02= & 4.42e-03= & 6.52E-03 \\
          & 200   & 9.56e-03= & 3.31e-02= & 3.53e-03= & 3.24e-02= & 8.55e-04= & 3.25e-02= & 4.79e-03= & 3.71e-03= & \textbf{8.65e-05} \\
          & 500   & 8.45e-03= & 1.19e-02= & \textbf{2.68e-06=} & 1.27e-02= & 1.35e-03= & 4.44e-01- & 3.21e-03= & 6.53e+00- & 3.71E-04 \\
          & 1000  & 4.49e-03= & 5.78e-03= & \textbf{1.83e-07=} & 5.20e-03= & 1.04e-03= & 3.92e+00- & 1.73e-03= & 2.70e+03- & 4.29E-04 \\
          & 2000  & 1.38e-03= & 1.27e-03= & 3.74e+00- & 3.57e+00- & 4.69e-04= & 2.56e+07- & 3.61e-04= & 4.11e+03- & \textbf{1.60e-04} \\
          & 5000  & 1.49e-04= & 1.26e-04= & 7.40e+03- & 7.32e+03- & 2.90e-06= & 1.84e+00- & \textbf{3.84e-08=} & 3.73e+03- & 1.15E-05 \\
    \midrule
    \multirow{6}[0]{*}{LSMOP5} & 100   & 6.87e-02= & 2.11e+01- & 5.73e+00- & 7.40e+01- & 1.02e-01- & 6.92e-01- & 5.20e-01- & 4.35e-02= & \textbf{3.73e-02} \\
          & 200   & 1.67e+00- & 8.43e+01- & 4.47e+00- & 2.70e+02- & 1.23e-01- & 1.00e+00- & 3.68e+00- & 8.70e-02- & \textbf{1.02e-02} \\
          & 500   & 1.43e+01- & 1.47e+02- & 4.87e-01- & 3.39e+02- & 1.10e-01- & 1.28e+04- & 6.81e+00- & 6.26e+00- & \textbf{1.52e-02} \\
          & 1000  & 6.96e+01- & 2.26e+02- & 3.99e-02= & 4.21e+02- & 1.38e-01- & 1.51e+04- & 8.03e+00- & 2.62e+03- & \textbf{1.83e-02} \\
          & 2000  & 2.59e+02- & 2.92e+02- & 3.08e+00- & 2.53e+00- & 2.19e-01- & 2.56e+07- & 9.26e+00- & 4.19e+03- & \textbf{1.50e-02} \\
          & 5000  & 3.83e+02- & 2.90e+02- & 7.36e+03- & 7.38e+03- & 2.17e-01- & 5.32e+03- & 9.14e+00- & 3.68e+03- & \textbf{1.03e-02} \\
    \midrule
    \multirow{6}[0]{*}{LSMOP6} & 100   & 7.48e-01- & 1.32e+06- & 6.84e+05- & 4.39e+07- & 9.01e-02= & 7.21e+02- & 3.89e+01- & 2.07e+00- & \textbf{5.58e-02} \\
          & 200   & 5.38e+00- & 1.72e+07- & 7.23e+05- & 6.37e+08- & 1.71e-01- & 4.28e+05- & 8.80e+02- & 5.20e+02- & \textbf{6.48e-02} \\
          & 500   & 6.90e+03- & 1.31e+08- & 2.49e+04- & 9.02e+08- & 2.48e-01- & 9.24e+10- & 3.50e+04- & 5.43e+00- & \textbf{1.17e-01} \\
          & 1000  & 3.04e+06- & 2.28e+08- & 1.43e+03- & 4.00e+02- & 8.12e-01- & 1.51e+11- & 8.91e+04- & 2.65e+03- & \textbf{4.44e-02} \\
          & 2000  & 3.83e+07- & 2.99e+08- & 2.26e+02- & 2.85e+00- & 1.56e+00- & 2.56e+07- & 8.36e+04- & 4.06e+03- & \textbf{4.82e-01} \\
          & 5000  & 2.41e+08- & 4.49e+08- & 7.40e+03- & 7.30e+03- & 4.83e-01- & 2.16e+09- & 1.05e+05- & 3.53e+03- & \textbf{1.74e-01} \\
    \midrule
    \multirow{6}[0]{*}{LSMOP7} & 100   & 8.61e-01- & 2.73e+00- & 4.84e+03- & 5.14e+00- & \textbf{1.47e-03=} & 1.18e-02= & 1.63e-01- & 5.61e-02= & 2.48E-02 \\
          & 200   & 5.05e-01- & 7.12e-01- & 2.33e+03- & 6.72e+05- & 8.31e-04= & 1.85e-01- & 2.08e-01- & 6.75e+02- & \textbf{6.27e-04} \\
          & 500   & 3.83e-01- & 4.25e-01- & 2.13e+01- & 9.87e+05- & 8.58e-02= & 7.89e+11- & 2.55e-01- & 1.01e+01- & \textbf{4.65e-02} \\
          & 1000  & 2.94e-01- & 2.93e-01- & 1.74e+00- & 3.95e+02- & 1.07e-01- & 3.38e+09- & 1.96e-01- & 2.70e+03- & \textbf{5.09e-02} \\
          & 2000  & 2.54e-01- & 2.54e-01- & 2.60e+00- & 1.41e+00- & 1.25e-01- & 2.56e+07- & 2.30e-01- & 4.02e+03- & \textbf{7.00e-02} \\
          & 5000  & 1.08e-01- & 1.08e-01- & 7.34e+03- & 7.32e+03- & 2.80e-02= & 4.16e+09- & 1.00e-01- & 3.61e+03- & \textbf{2.07e-02} \\
    \midrule
    \multirow{6}[0]{*}{LSMOP8} & 100   & 5.02e-02= & 6.75e-01- & 4.22e-01- & 5.29e-02= & 2.23e-02= & 6.77e-01- & 3.13e-01- & 3.80e-02= & \textbf{1.34e-02} \\
          & 200   & 7.37e-02- & 5.88e-01- & 2.81e-01- & 4.19e-01- & 5.59e-02- & 7.85e-01- & 2.85e-01- & 6.84e+02- & \textbf{3.54e-04} \\
          & 500   & 3.82e-01- & 6.27e-01- & 1.42e-01- & 3.36e-01- & 2.13e-02= & 3.17e+03- & 2.30e-01- & 7.41e+00- & \textbf{1.78e-04} \\
          & 1000  & 6.81e-01- & 8.02e-01- & 2.43e-02= & 4.19e+02- & 3.00e-02= & 5.27e+03- & 2.26e-01- & 2.74e+03- & \textbf{2.82e-04} \\
          & 2000  & 7.36e-01- & 7.87e-01- & 1.68e+00- & 3.83e+00- & 4.96e-02= & 2.56e+07- & 2.22e-01- & 4.22e+03- & \textbf{4.37e-04} \\
          & 5000  & 7.87e-01- & 7.17e-01- & 7.36e+03- & 7.55e+03- & 4.25e-02= & 5.06e+00- & 2.13e-01- & 3.65e+03- & \textbf{8.35e-05} \\
    \midrule
    \multirow{6}[0]{*}{LSMOP9} & 100   & 2.84e-01+ & 9.39e+02- & 3.48e+02- & 4.04e-01+ & 3.69e-01+ & 1.53e+02- & \textbf{1.48e-01+} & 3.65e-01+ & 9.59E-01 \\
          & 200   & 2.60e+00- & 1.98e+03- & 2.13e+02- & 1.36e+04- & \textbf{6.95e-03+} & 4.77e+02- & 3.64e-01- & 6.38e+02- & 1.54E-01 \\
          & 500   & 5.31e+00- & 4.69e+03- & 7.91e+00- & 1.73e+04- & \textbf{4.01e-01+} & 1.33e+05- & 1.39e+00- & 7.85e+00- & 7.39E-01 \\
          & 1000  & 4.36e+01- & 6.76e+03- & \textbf{1.36e-03+} & 4.04e+02- & 2.10e-01= & 1.05e+05- & 9.09e+02- & 2.65e+03- & 2.04E-01 \\
          & 2000  & 2.09e+02- & 6.69e+03- & 2.72e+00- & 2.55e+00- & 9.48e-01- & 2.56e+07- & 3.84e+03- & 4.21e+03- & \textbf{8.74e-01} \\
          & 5000  & 6.73e+02- & 8.46e+03- & 7.20e+03- & 7.24e+03- & 3.14e-02= & 4.50e+04- & 2.78e+03- & 3.59e+03- & \textbf{4.39e-03} \\
    \midrule
    \multicolumn{2}{c}{(+/-/=)} & {1/38/15} & {0/43/11} & {1/41/12} & {1/44/9} & {3/16/35} & {0/49/5} & {1/41/12} & {1/44/9} &  \\
    \bottomrule
    \end{tabular}%
  \label{tab:igd_insensitivity}%
\end{table*}%

\subsection{General Performance}
\label{exp:gp}
Due to length limitations, the statistical results of the average IGD and HV values over 20 runs are placed in the Table S-II and S-III of the supplementary materials, respectively. Generally, LMOMCTS has outperformed all seven state-of-the-art large-scale MOEAs (not counting NSGA-II as a baseline).
\par
LMOMCTS obtains 30 best results in 54 test cases, outperforming the other eight compared methods. WOF achieves the 9 best results. MOEA/DVA, LMEA, and LMOCSO obtain the 5 best results. Other algorithms do not obtain the best result. The proposed LMOMCTS is also better than those of the other eight comparative algorithms with the indicator of HV.
\par
The proposed algorithm performs slightly worse on LSMOP9. Based on the characteristics of the benchmark \cite{7553457}, we believe that the proposed algorithm is more advantageous in solving the LSMOP with linear PF and convex PF. When the PF is disconnected, the problem transformation-based method (e.g., WOF) has an edge in performance. Nevertheless, the proposed algorithm uses the Monte Carlo tree search to sample different decision variables on different nodes for optimization so that the algorithm still delivers fairly acceptable results on these test instances.
\par
The experimental results show that LMOMCTS is a method with good convergence and diversity for most test problems. The proposed LMOMCTS can obtain competitive performance on the LSMOP test problems, which is the basis of applying large-scale MOEAs to solve real-world problems.
\par
\begin{figure*}[htbp]
    \centering
    \subfloat[Nondominated solutions obtained by LMOCSO.]{\includegraphics[width=0.3\hsize]{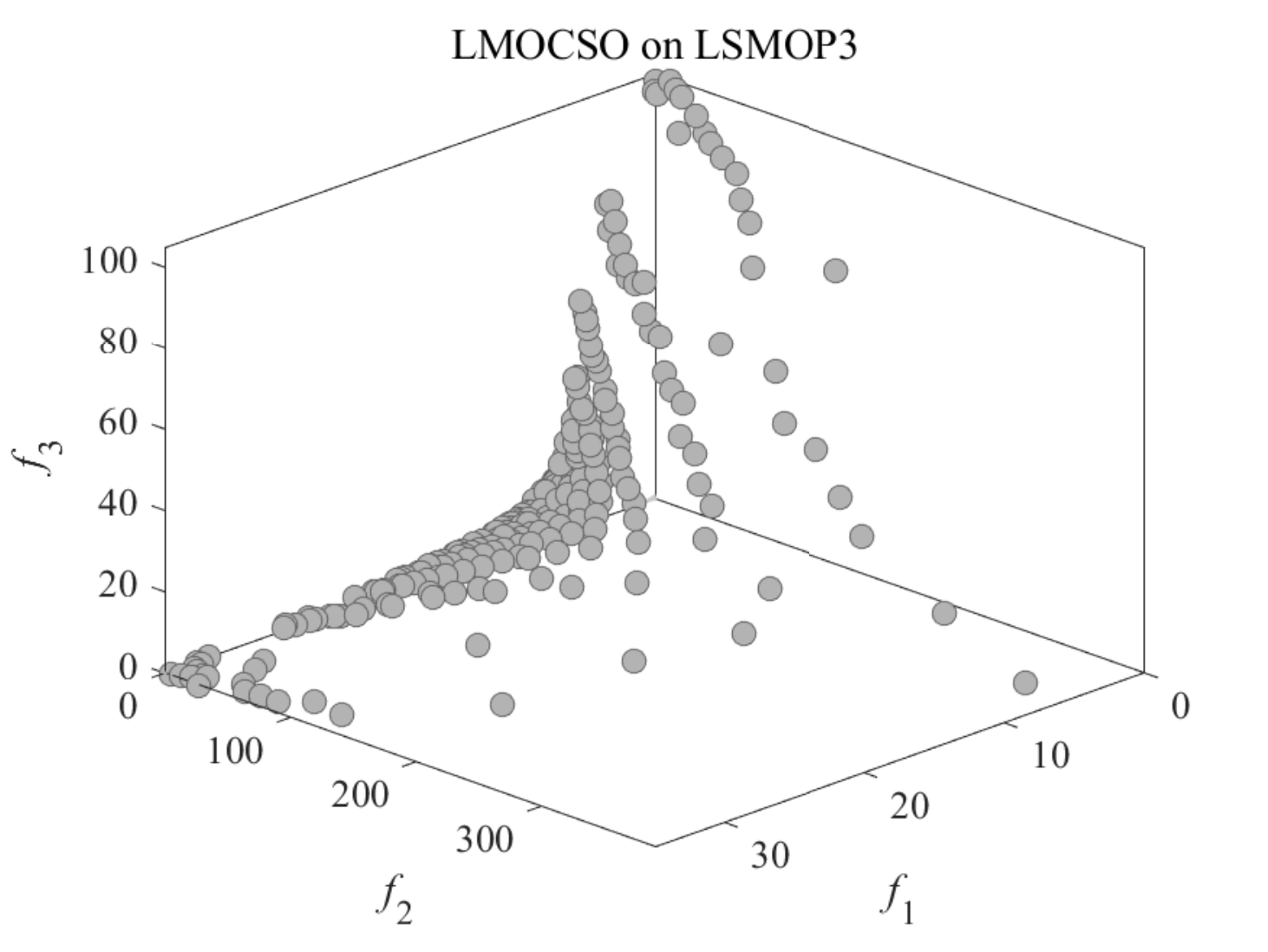}}\hspace{3mm}
    \subfloat[Nondominated solutions obtained by WOF.]{\includegraphics[width=0.3\hsize]{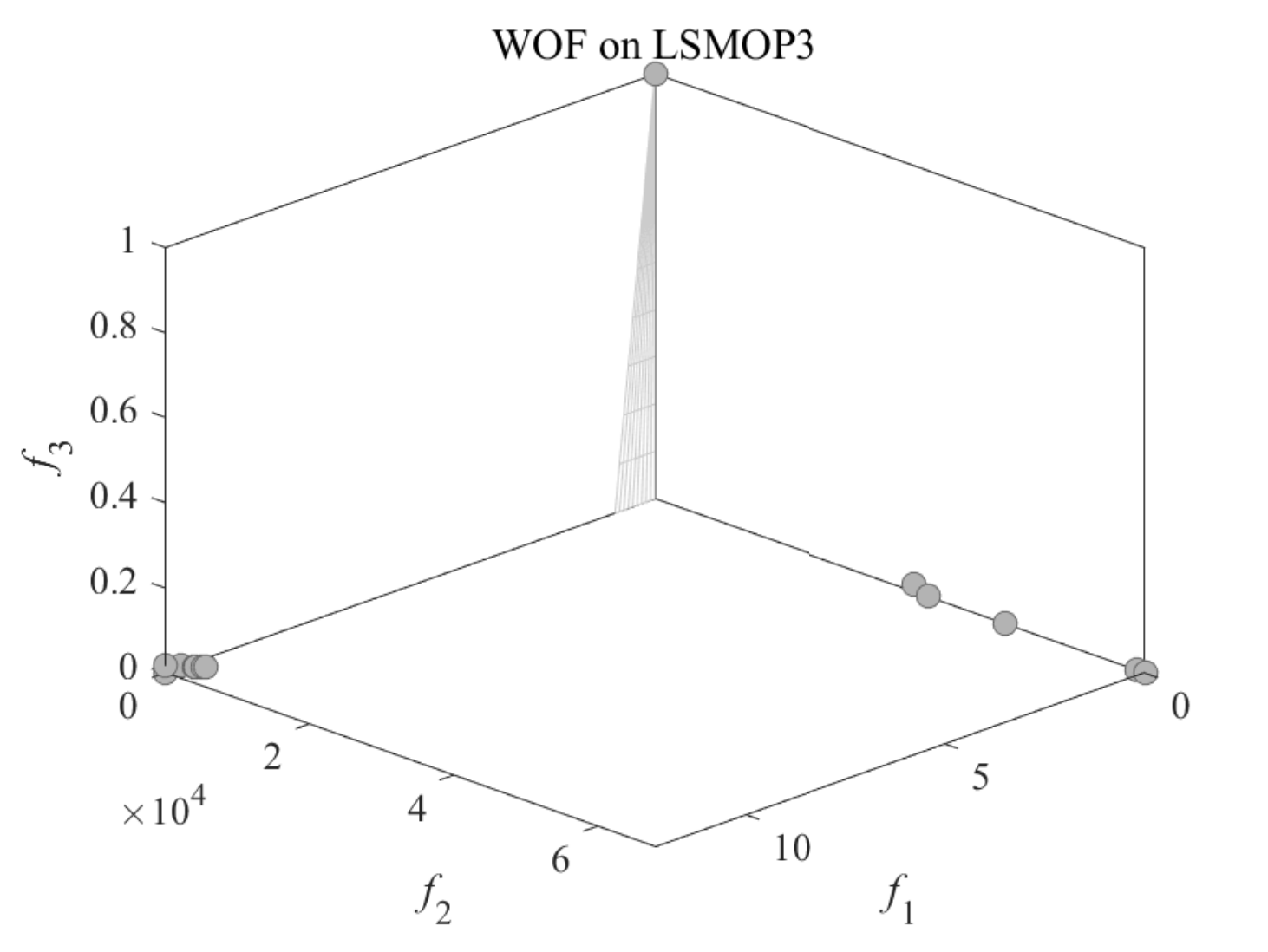}}\hspace{3mm}
    \subfloat[Nondominated solutions obtained by LMOMCTS.]{\includegraphics[width=0.3\hsize]{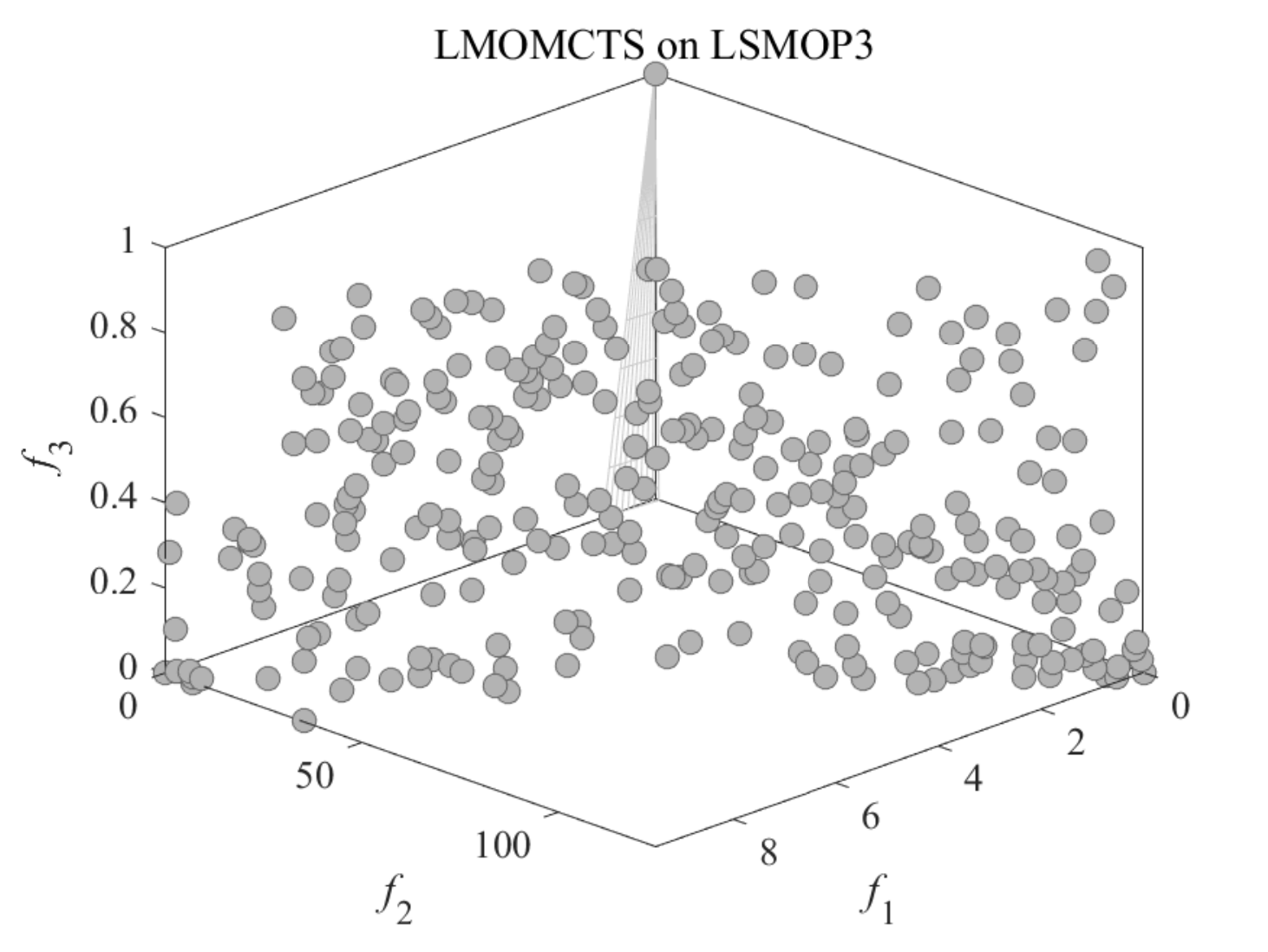}}
    \caption{Nondominated solutions obtained by compared methods on tri-objective LSMOP3 with 1000 decision variables. Please refer to Figs. S-1 to S-3 of the supplementary materials for other algorithms.}
    \label{fig:pof}
\end{figure*}

{Nondominated solutions obtained by the compared methods on tri-objective LSMOP3 with 1000 decision variables are presented in Fig. \ref{fig:pof}, where the mesh is the real PF. From the distribution of these nondominated solutions in the target space, it can be seen that although none of these algorithms perform well on the LSMOP3 problem, only our algorithm's output is closer to PF; that is, the output points are closer to the grid. Second, the output of the algorithm has higher diversity, i.e., more populations are near PF.}

\subsection{Performance Insensitivity Study}
We present the insensitive-IGD of all competing algorithms in Table \ref{tab:igd_insensitivity}, and the results of the insensitive-HV are given in the Table S-IV of the supplementary materials. Overall, the values of insensitive-IGD and insensitive-HV of our algorithm remarkably outperformed other algorithms, validating that the designed method improves the performance insensitivity of the proposed algorithm.
\par
Our algorithm achieves 34 best results in total for the indicator of insensitive-IGD, which means the proposed algorithm runs 34 times on 54 instances with better performance insensitivity than that of other algorithms. Aside from obtaining the highest number of minimum variances, a notable point is that our algorithm does not obtain an insensitive-IGD greater than one on any given test instance, while the other compared algorithms have instances with an insensitive-IGD greater than one. This means that even if our algorithm does not achieve the best results on all tests, our algorithm outputs results that are very close to the results of the best algorithm.
\par
Combined with the discussion in section \ref{exp:gp}, our algorithm has an advantage in performance and insensitivity, while other algorithms are more sensitive. The insensitive-IGD and insensitive-HV values of our algorithm on all test instances are maintained at a small order of magnitude, ensuring the performance insensitivity of the algorithm in practical applications.
\par
To avoid type-I error (i.e., the null hypothesis is wrongly rejected) caused by repeated pair-based hypothesis testing with the same dataset\cite{20.500.11850/84395}, we also conducted experiments on the WFG benchmark\cite{1705400}. The experimental results are given in the Tables S-V and S-VI of the supplementary materials. Similarly, our algorithm obtains not only the best mean results on the WFG suite but also the best insensitive-IGD results.
\par
{One motivation of the proposed method is to improve the performance insensitivity of the algorithm when solving real-world problems. We present experimental results on a set of real-world LSMOPs, TREE, \cite{8962275} and presented in Tables \ref{tab:TREE_i_igd}. On the TREE test set, our method obtains not only the best mean results (5 out of 6) but also the best insensitive-IGD results (5 out of 6). Full experimental results are given in the Tables S-VII S-VIII, S-IX, and S-X of the supplementary materials.}
\par
\begin{figure}[htbp]
    \centering
    \subfloat[{LMOMCTS Solves LSMOP3}]{\includegraphics[width=0.48\hsize]{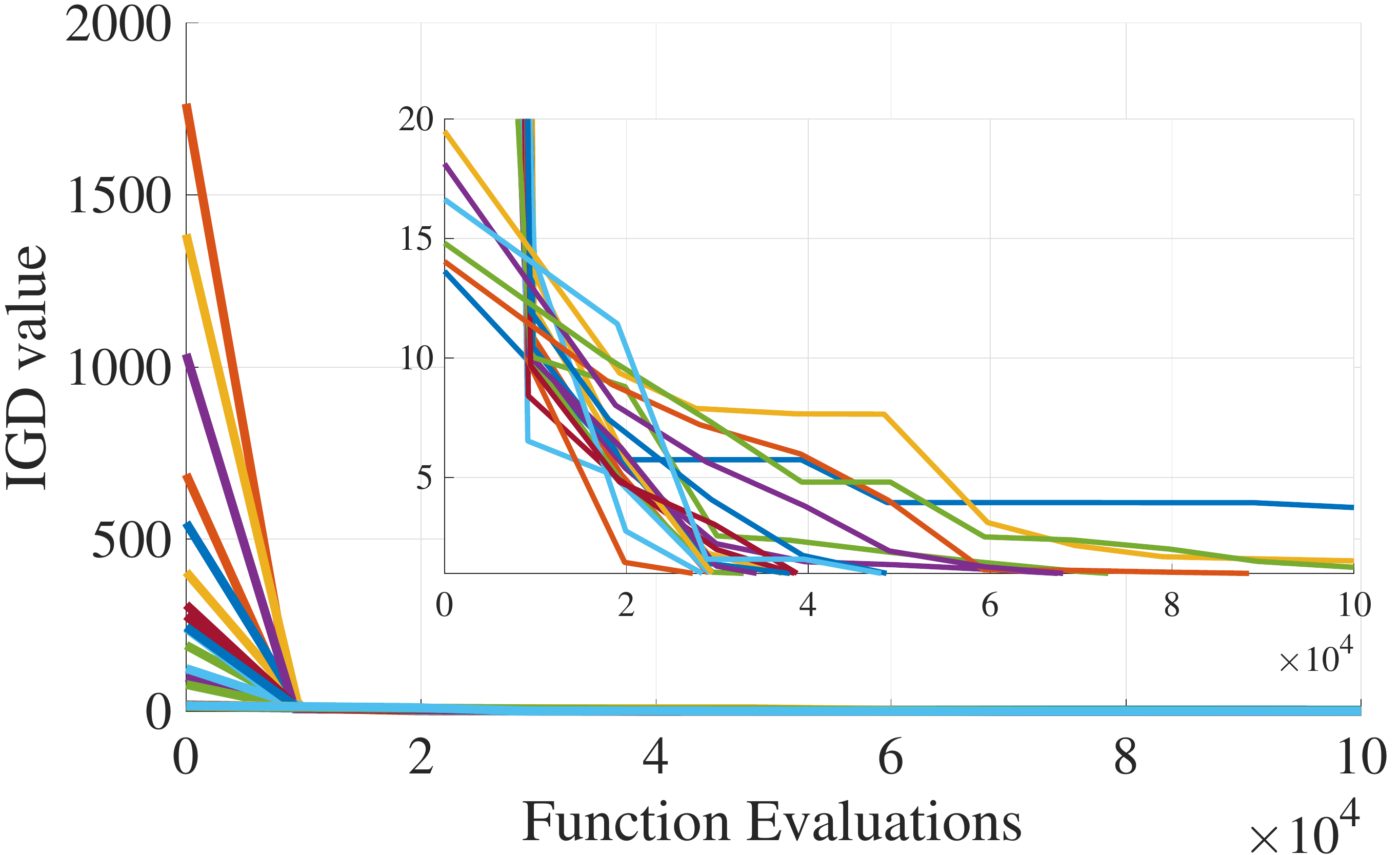}}
    \subfloat[{LMOMCTS Solves LSMOP6}]{\includegraphics[width=0.48\hsize]{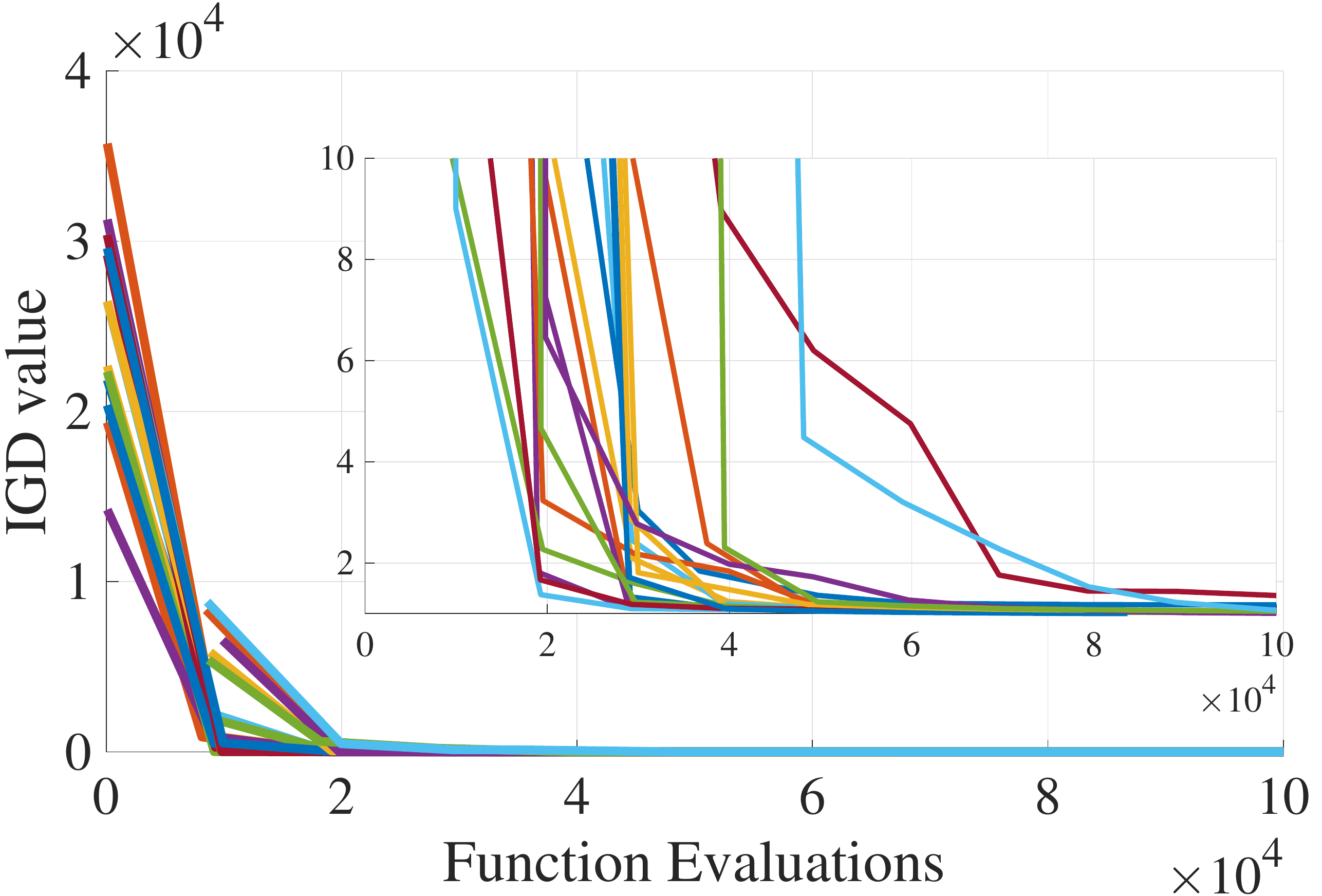}}
    \caption{{Convergence profiles of the IGD values obtained by LMOMCTS on LSMOP3 for 20 times with different initial population. Please refer to Figs. S-4 to S-6 of the supplementary materials for all results.}}
    \label{fig:insensitive_d}
\end{figure}
\begin{figure}[htbp]
    \centering
    \subfloat[{LMOMCTS Solves LSMOP3.}]{\includegraphics[width=0.48\hsize]{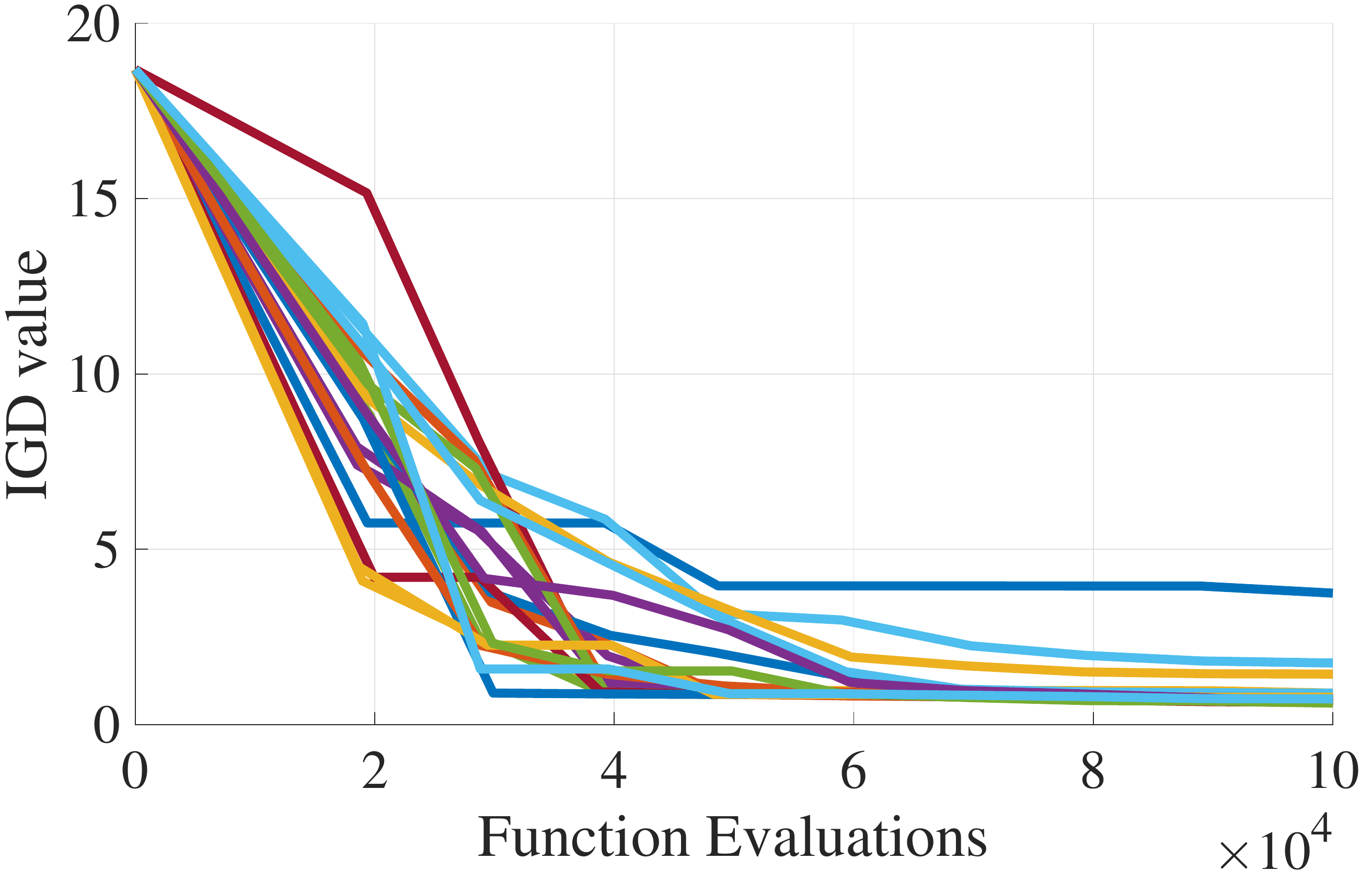}}
    \subfloat[{LMOMCTS Solves LSMOP6}]{\includegraphics[width=0.48\hsize]{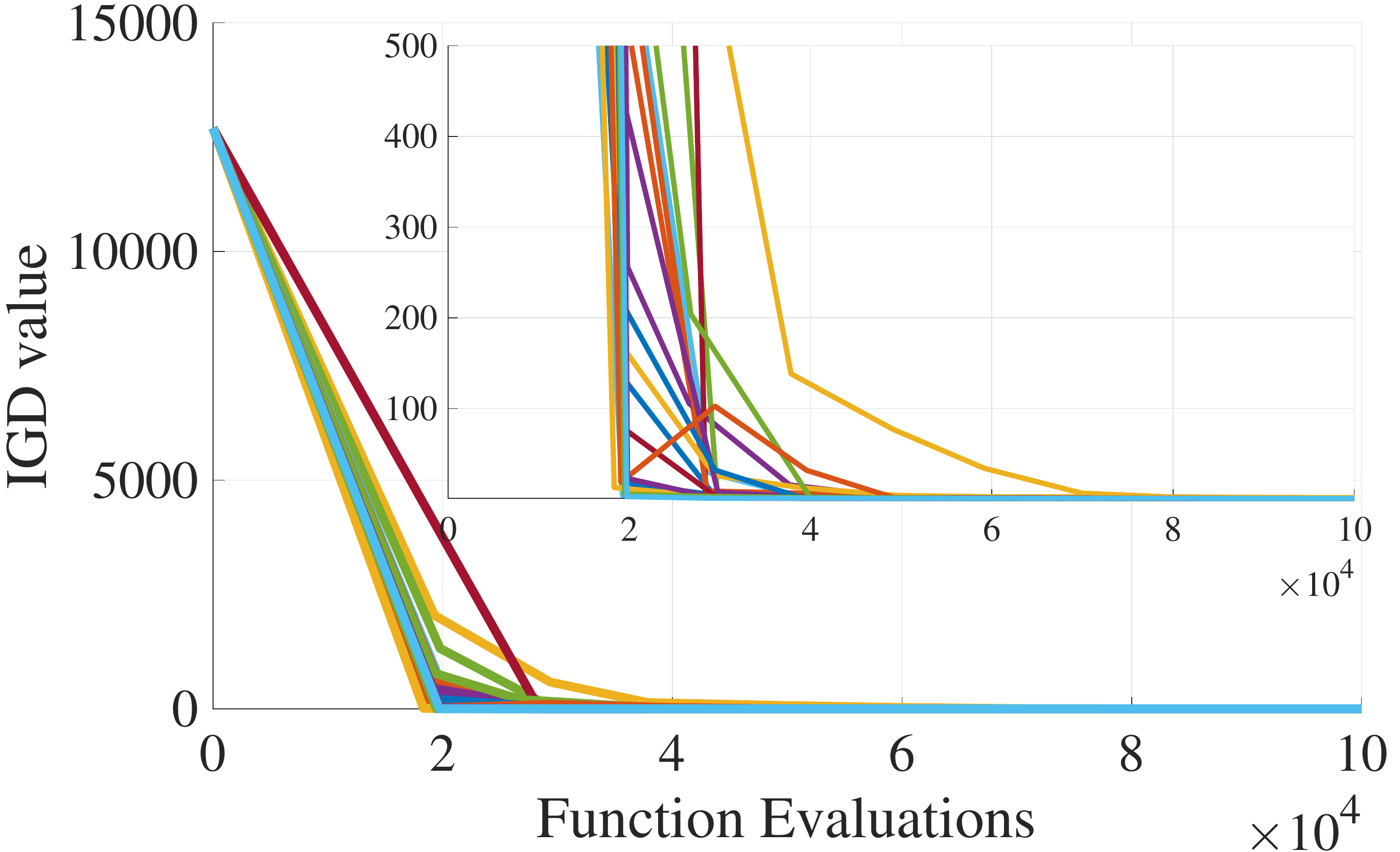}}
    \caption{{Convergence profiles of the IGD values obtained by LMOMCTS on LSMOP6 for 20 times with same initial population. Please refer to Figs. S-7 to S-9 of the supplementary materials for all results.}}
    \label{fig:insensitive_s}
\end{figure}

\begin{table*}[htbp]
\scriptsize
  \centering
  \color{DeepRed}
  \caption{Insensitive-IGD Values Obtained By 9 Compared Algorithms on 6 Real-world Instances From TREE Test Suite. The Best Result in Each Row is Highlighted in Bold. Results that took more than 50,000s are represented by '-', and results that are not obtained before EFs are exhausted are represented by 'NaN'}
  \begin{tabular}{cccccccccccc}
    \toprule
    Problem & D     & FE    & NSGA-II & CCGDE3 & MOEA/DVA & LMEA  & WOF   & S3-CMA-ES & LMOCSO & GMOEA & LMOMCTS \\
    \midrule
    TREE1 & 6.0E+03 & 2.0E+05 & 3.51e+04- & NaN-  & -  & -  & 2.87e+04- & -  & 3.84e+04- & -  & \textbf{1.04e+04} \\
    \midrule
    TREE2 & 6.0E+03 & 4.0E+05 & 7.37e+05- & NaN-  & -  & -  & 6.89e+05- & -  & 7.75e+05- & -  & \textbf{2.45e+03} \\
    \midrule
    TREE3 & 6.0E+03 & 2.0E+05 & NaN-  & NaN-  & -  & -  & 1.01e+04- & -  & NaN-  & -  & \textbf{4.18e+02} \\
    \midrule
    TREE4 & 1.2E+04 & 1.0E+06 & 3.10e+04- & NaN-  & -  & -  & 3.83e+04- & -  & 3.84e+04- & -  & \textbf{1.48e+03} \\
    \midrule
    TREE5 & 6.0E+04 & 6.0E+05 & 2.86e+00- & NaN-  & -  & -  & 7.25e+05- & -  & NaN-  & -  & \textbf{1.35e+03} \\
    \midrule
    TREE6 & 1.2E+04 & 4.0E+05 & -  & NaN=  & -  & -  & -  & -  & -  & -  & NaN \\
    \midrule
    \multicolumn{3}{c}{(+/-/=)} & {0/5/1} & {0/5/1} & {0/5/1} & {0/5/1} & {0/5/1} & {0/5/1} & {0/5/1} & {0/5/1} &  \\
    \bottomrule
    \end{tabular}%
  \label{tab:TREE_i_igd}%
\end{table*}%
Similar to the preliminary experiments conducted in the Introduction section, we present the results of running the proposed algorithm 20 times under the same environment. Specifically, Figs. \ref{fig:insensitive_d}. (a) and (b) are the result of running LMOMCTS on LSMOP3 and LSMOP6 under 20 different initial populations, and Figs. \ref{fig:insensitive_s}. (a) and (b) are the result of LMOMCTS running 20 times with the same initial population, all of which reuse the population used in the introduction section. 
\par
It can be seen from the figures that the optimization results of our algorithm across multiple runs are kept in a stable and good range, and there will be no apparent fluctuations in multiple runs. It is worth noting that when solving LSMOP6 with the same initial population, the result of our algorithm in the early stage is not ideal in a certain run, but it can still obtain a better convergence result at the end, which ensures the insensitivity of the proposed algorithm.
\par
\begin{figure}[htbp]
    \centering
    \subfloat{\includegraphics[width=0.72\hsize]{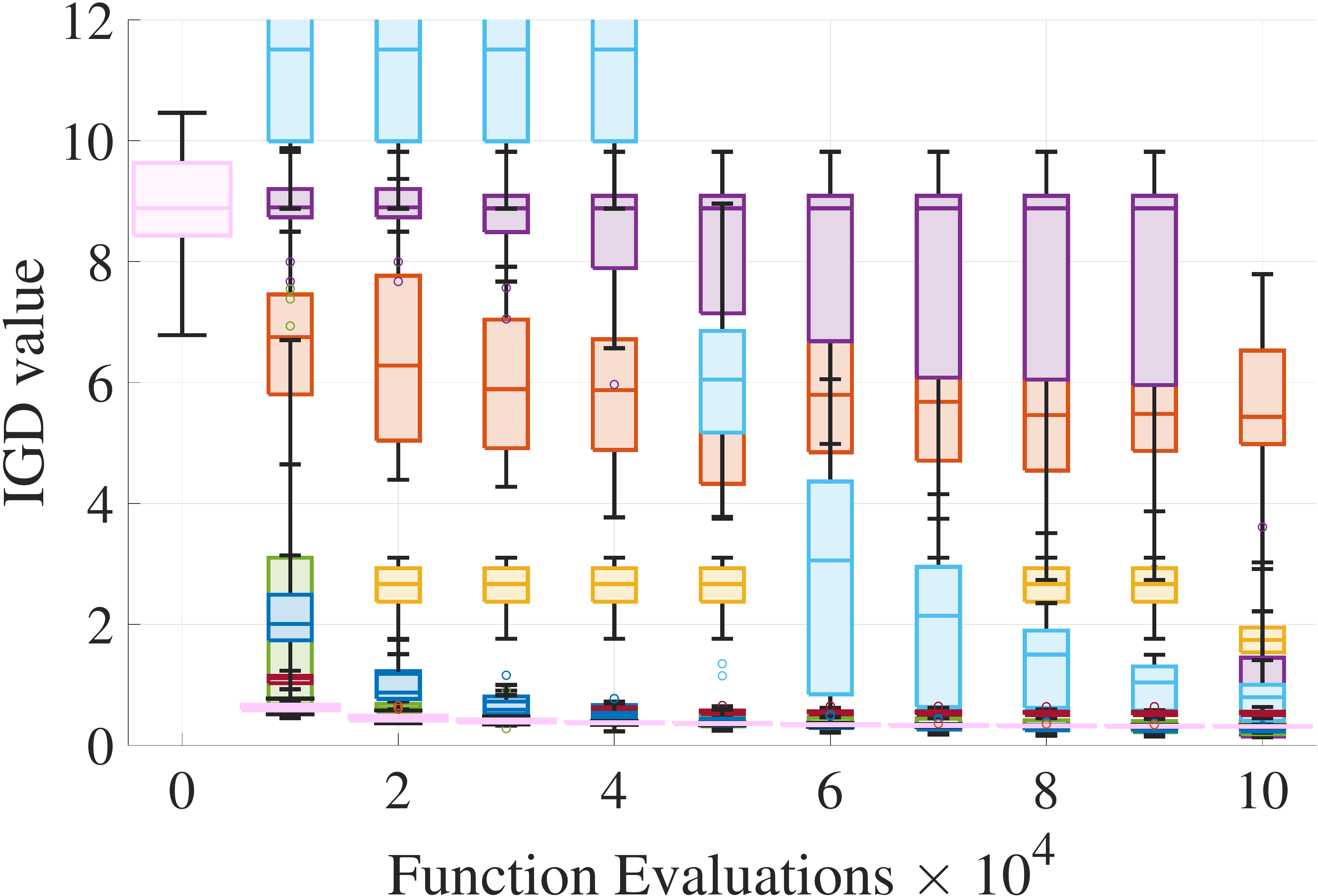}}
    \subfloat{
        \raisebox{0.45\height}
        {\includegraphics[width=0.18\hsize]{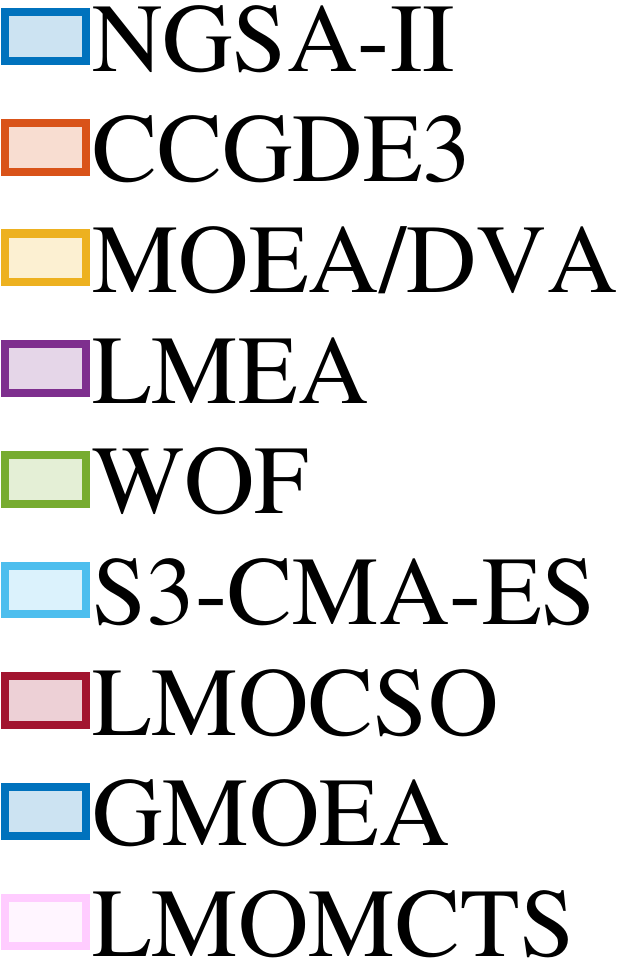}}
    }
    \caption{Box plot of the IGD values obtained by all compared algorithms on LSMOP1 20 times with different initial populations. Please refer to Figs. S-10 to S-12 of the supplementary materials for results on other problems.}
    \label{fig:insensitive}
\end{figure}

To intuitively demonstrate the advantages of the proposed algorithm's performance insensitivity in solving the LSMOP, we show the box plot of IGD values obtained by all the comparison algorithms over 20 runs in Fig. \ref{fig:insensitive}. As seen from the figure, our algorithm not only outputs better solutions but also ensures the insensitivity of the optimization results over multiple runs.
\par
Analyses of other problems are presented in the supplementary materials, and our algorithm performs similarly on other benchmarks.
\subsection{Convergence and Computational Efficiency}
\begin{figure*}[htbp]
    \centering
    \subfloat[Variations in IGD values achieved by compared algorithms on tri-objective LSMOP1 with 100 decision variables.]{\includegraphics[width=0.26\hsize]{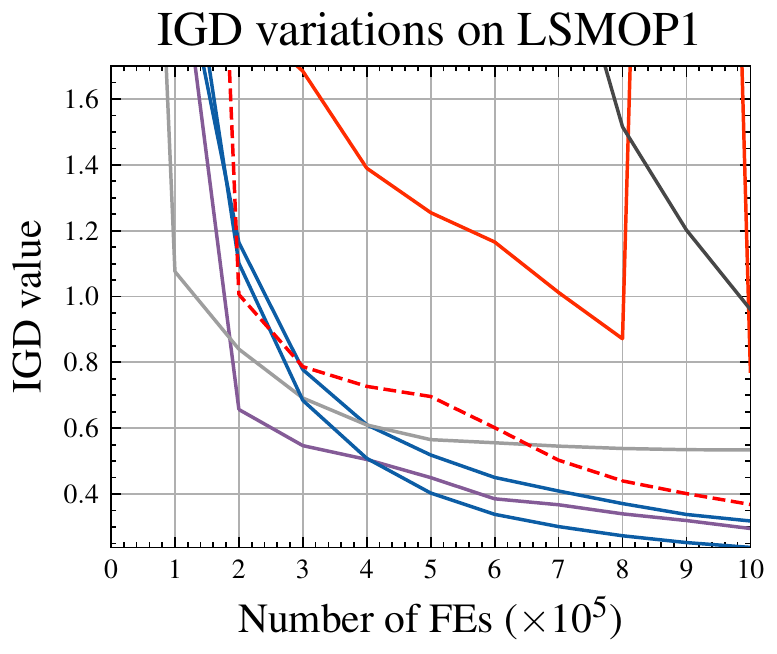}}\hspace{3mm}
    \subfloat[Variations in HV values achieved by compared algorithms on tri-objective LSMOP1 with 100 decision variables.]{\includegraphics[width=0.26\hsize]{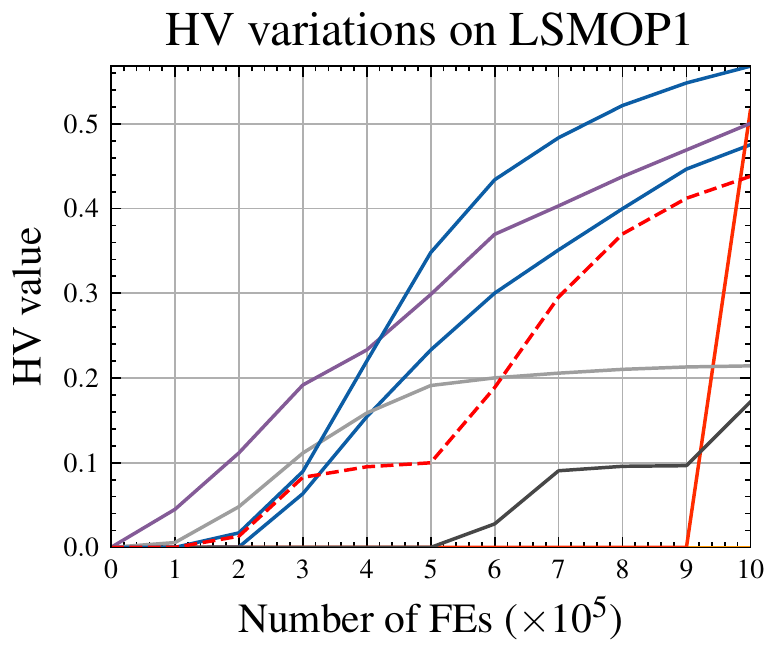}}\hspace{3mm}
    \subfloat[Average computation time of compared algorithms on tri-objective LSMOP1.]{\includegraphics[width=0.39\hsize]{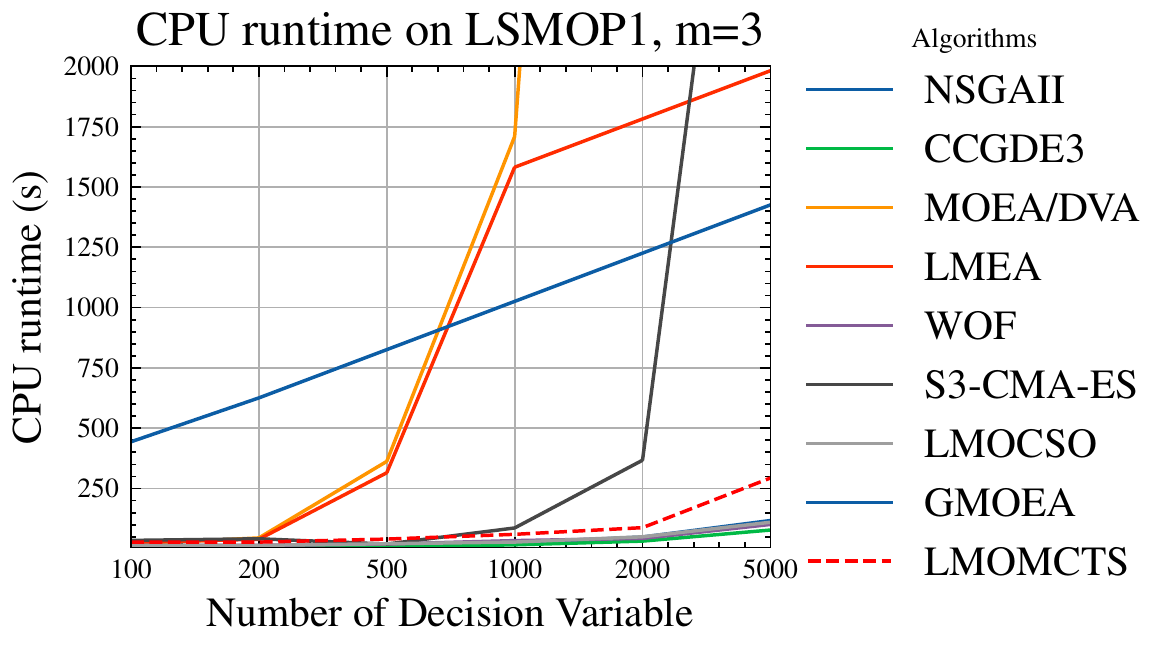}}\hspace{3mm}
    \caption{Analysis on Convergence and Computational Efficiency}
    \label{fig:ACCE}
\end{figure*}

Another important aspect of performance insensitivity is to maintain a good result when the algorithm stops in unanticipated situations. {The variations in IGD values achieved by all compared algorithms on the tri-LSMOP1 problem are presented in Fig. \ref{fig:ACCE}. (a). From the figure, the IGD value of our algorithm is not the best at all times, but the IGD value decreases monotonically during the running of the algorithm. It can be seen that our algorithm is not sensitive to this parameter. As long as e is selected in an appropriate interval, our proposed method has good performance.} Similarly, the variations in HV values achieved by the compared algorithms on LSMOP1 are given in Fig. \ref{fig:ACCE}. (b), which illustrates that our algorithm achieved monotonic (the performance is not degraded) results.
\par
{In addition to the performance insensitivity, the algorithm's computational efficiency is also critical in practical applications. We present the average running time of all the compared algorithms on LSMOP1-LSMOP3 with decision variables varying from 100 to 5,000. As shown in Fig. \ref{fig:ACCE}. (c), our proposed LMOMCTS has a relatively lower computation time. The experimental results are consistent with the analysis of time complexity, our algorithm maintains relatively low time complexity under different dimensions.}
\par
\begin{figure}[htbp]
    \centering
    \subfloat[{Memory usage of compared algorithms during solving tri-
objective LSMOP1.}]{\includegraphics[width=0.45\hsize]{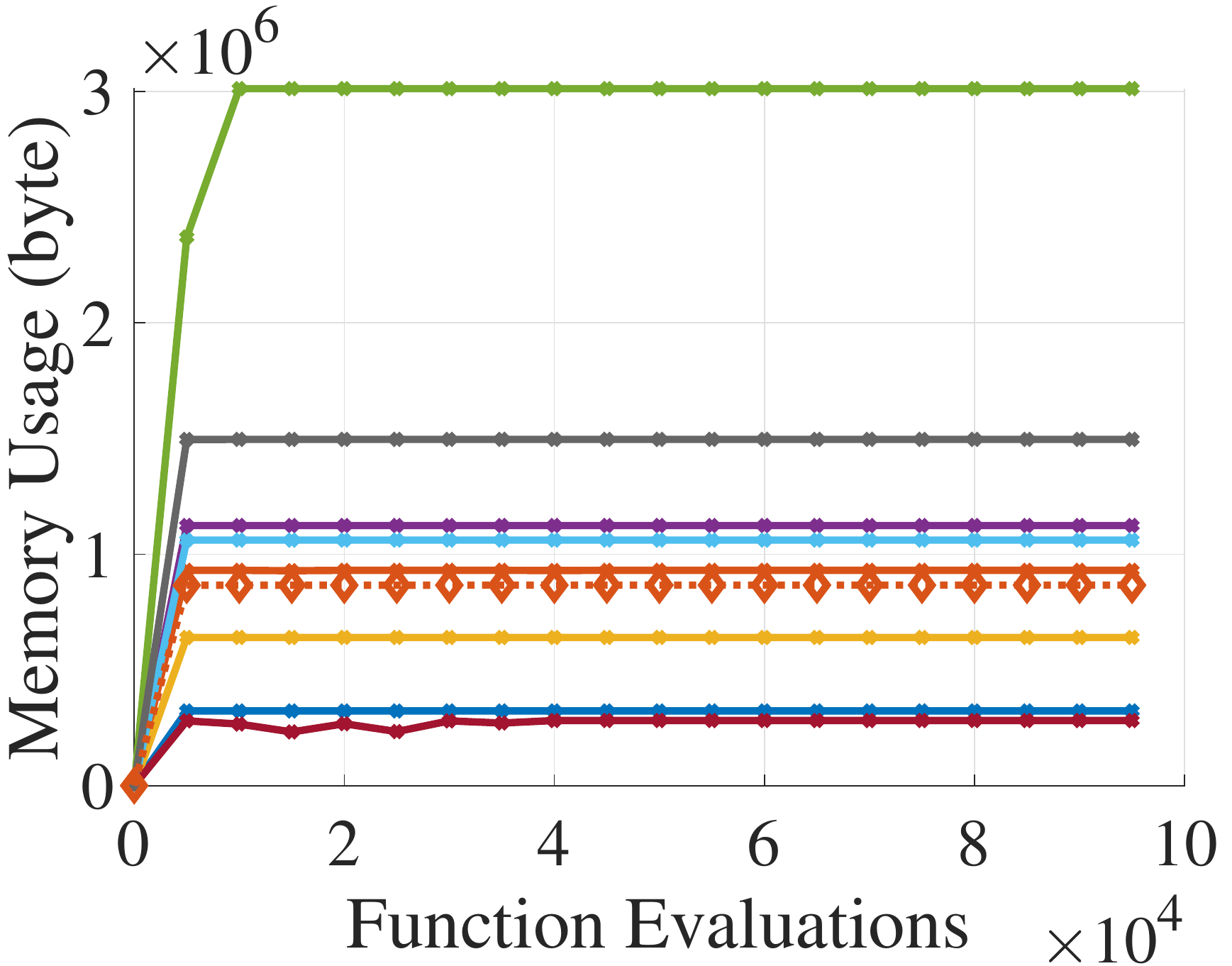}}
\hspace{3mm}
    \subfloat[{Peak memory usage of compared algorithms on LSMOP1 with different decision variable}]{\includegraphics[width=0.48\hsize]{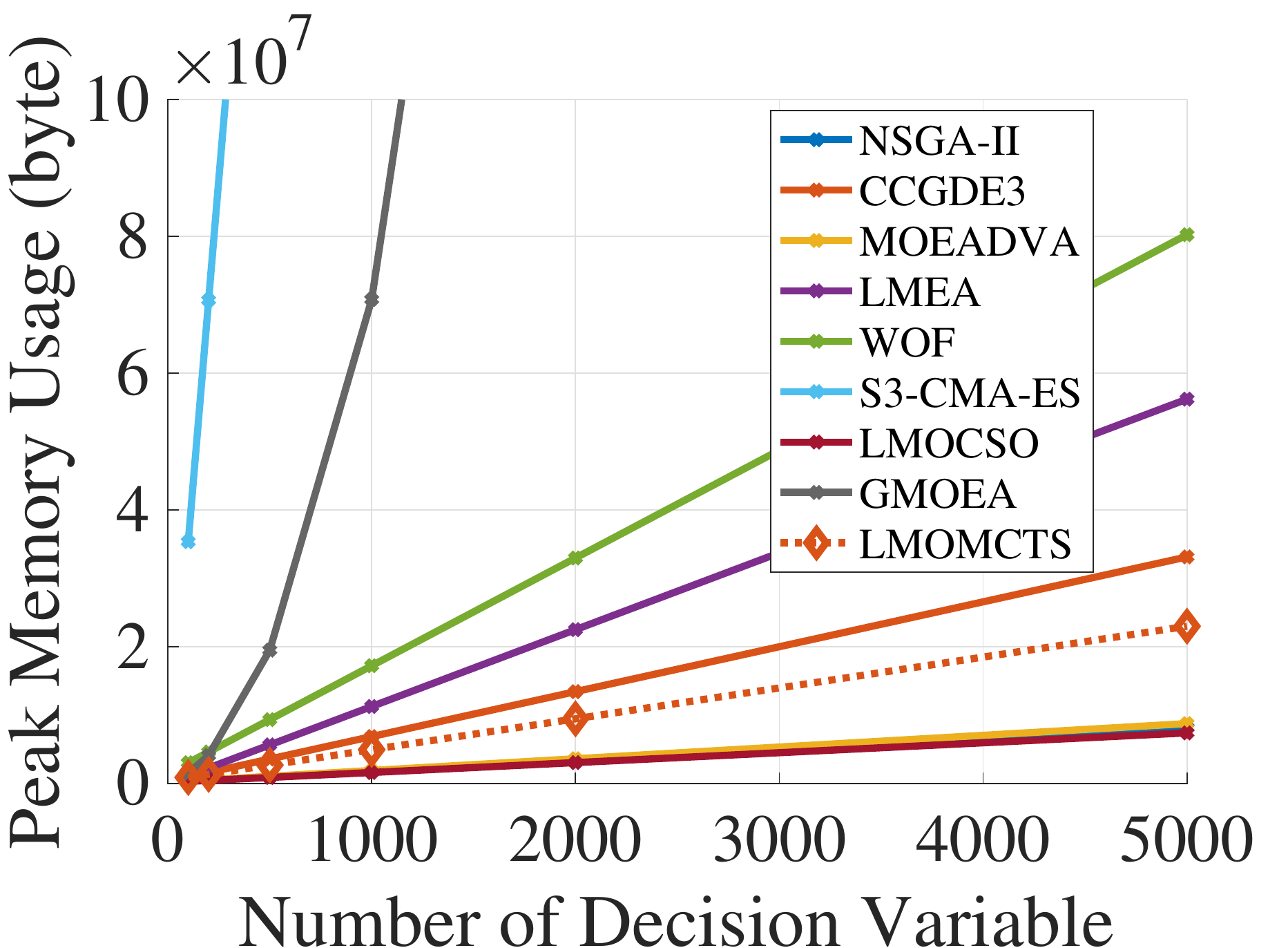}}
    \caption{{Memory usage of compared algorithms on tri-LSMOP1.}}
    \label{fig:space}
\end{figure}

{Besides the runtime complexity, we present the memory usage of all the compared algorithms during solving tri-objective LSMOP1 and the peak memory usage from decision variables 100 to 5,000 in Figs.~\ref{fig:space}. (a) and (b), respectively. First, based on the design of discarding nodes, the memory overhead does not increase continuously while running. Second, the memory usage of our algorithm does not explode when the dimension increases.}
\par
Figures on other problems are presented in the Figs. S-13 to S-21 of the supplementary materials. Our algorithm also shows consistent results on these problems.

\subsection{Parameter Sensitivity Analysis}
The key parameter of our proposed LMOMCTS is the ratio of sampling decision variables $f$, which is used to control the sampling size of decision variables. To analyze the effect of the parameter on the performance of the proposed algorithm, we conduct experiments on a set of LSMOP problems with 1,000 decision variables. {The parameter $f$ is set to 0.1, 0.2, 0.25, and 0.5, and the corresponding sampling sizes of the decision variables are 100, 500, 250, and 500, respectively. The experimental results are shown in Table \ref{tab:parameter}}.
\par
{The statistical results show that different settings of $f$ do not significantly affect the performance of the proposed algorithm. This phenomenon can be attributed to the design of the number of child nodes.} When the value of $f$ is small, that is, when the number of sampled decision variables is small, the number of branches $k$ calculated according to Eq. (\ref{equ:factork}) will grow larger so that the number of child nodes will increase accordingly, thereby ensuring that enough decision variables are optimized. Conversely, when $f$ becomes larger, $k$ decreases, allowing enough decision variables to be sampled.
\par
{Parameter $e$ controls the number of nodes that the algorithm would generate within the total function evaluations. A smaller $e$ means more nodes, and a larger $e$ means that the parent node will evolve deeper to generate a child node, which is also an adaptive parameter. We give the experimental results in the Table XI of the supplementary materials, and this parameter is also insensitive. It can be seen that our algorithm is not sensitive to this parameter. As long as $e$ is selected in an appropriate interval, our proposed method has good performance.}

\subsection{Discussion}
We conduct a series of experiments to test the IGD, HV, and insensitivity of the proposed method and analyze the convergence and computational efficiency of the algorithm. See supplementary materials for complete experiments\footnote{Since the journal does not allow supplementary materials to be attached to the new submission, the supplementary materials can be downloaded from the following address: \href{https://www.jianguoyun.com/p/DRL64ucQrLOVCxjt6PEEIAA}{Supplementary materials}}. We can see that our proposed algorithm has a better overall performance than representative state-of-the-art large-scale multiobjective optimization algorithms. On this basis, our algorithm also has a significant advantage in performance insensitivity.
\par
{First, the designed decision variable sampling for optimization provides the basis for dealing with large-scale decision variables. It enables the proposed method to output good optimization results, which is the essence of applying algorithms to practical problems. }
\par
{Furthermore, it can be seen from the experiments that some algorithms have good results in a few runs but still output poor results in many runs (Fig. \ref{fig:insensitive}). In this regard, we design two indicators, insensitive-IGD and insensitive-HV, to measure the performance insensitivity of the compared algorithms. Our algorithm produces multiple nodes in the early search stage, in which high variance may appear in the early stage in Fig. \ref{fig:insensitive}. Nevertheless, our algorithm would sooner evaluate the nodes and select promising nodes to search and produce robust optimization results in each run.}
\par
In summary, the experiments validate that the use of decision variable sampling for optimization and MCTS in solving LSMOP provides optimization a unique advantage. The proposed algorithm can sample different decision variables to generate different child nodes and evaluate them and select the most promising nodes for optimization according to the evaluation results, thereby ensuring the algorithm's performance and improving the insensitivity of the optimization results.

\begin{table}[htbp]
  \centering
  \caption{Average IGD Values Achieved by LMOMCTS With Different Settings of Ratio of Sampling Decision Variables $f$}
    \begin{tabular}{cccccc}
    \toprule
    Problem & D     & $f=0.1$ & $f=0.2$ & $f=0.25$ & $f=0.5$ \\
    \midrule
    LSMOP1 & 1000  & 7.69e-01 & 5.92e-01 & 6.41e-01 & 5.50e-01 \\
    \midrule
    LSMOP2 & 1000  & 3.77e-02 & 3.70e-02 & 3.73e-02 & 3.62e-02 \\
    \midrule
    LSMOP3 & 1000  & 3.21e+00 & 9.72e-01 & 8.61e-01 & 8.61e-01 \\
    \midrule
    LSMOP4 & 1000  & 1.09e-01 & 9.81e-02 & 9.92e-02 & 9.47e-02 \\
    \midrule
    LSMOP5 & 1000  & 2.07e+00 & 5.38e-01 & 5.39e-01 & 5.41e-01 \\
    \midrule
    LSMOP6 & 1000  & 3.25e+01 & 1.53e+00 & 1.40e+00 & 1.68e+00 \\
    \midrule
    LSMOP7 & 1000  & 1.05e+00 & 9.61e-01 & 8.75e-01 & 9.42e-01 \\
    \midrule
    LSMOP8 & 1000  & 5.89e-01 & 3.07e-01 & 3.58e-01 & 2.79e-01 \\
    \midrule
    LSMOP9 & 1000  & 3.55e+00 & 1.15e+00 & 1.15e+00 & 1.15e+00 \\
    \bottomrule
    \end{tabular}%
  \label{tab:parameter}%
\end{table}%
\section{Conclusion and Future works}
\label{sec:con}
This paper proposes an algorithm based on Monte Carlo tree search, called LMOMCTS, to improve the performance insensitivity when solving LSMOPs. Based on the MCTS, the search is not limited to one population, and the evaluation on the node guides the algorithms to avoid searching on underperforming populations. With the design of decision variable sampling for optimization, the algorithm can reduce the dimensionality and search on different decision variables. To measure the algorithm's sensitivity, we propose two new metrics that can simultaneously characterize the performance of the optimization results and the consistency of multiple runs. Experimental results show that, compared with several latest large-scale MOEAs, the proposed LMOMCTS can improve the performance insensitivity while ensuring the performance in solving LSMOPs.
\par
This paper presents a preliminary work exploiting MCTS to improve the performance insensitivity of LSMOPs. Many significant problems can be modeled as LSMOP, and in practice, the algorithm requires good performance insensitivity. Future work may take several possible directions. First, some details of this method can be improved. For example, other methods can be used to generate offspring populations from parent nodes. Second, we will consider combining MCTS with other evolutionary algorithms. In addition, due to large-scale decision variables, we will try to introduce some machine learning approaches \cite{9097186} \cite{9199822} into the analysis and the calculation for LSMOPs.

\section*{Acknowledgments}
This work was supported in part by the National Natural Science Foundation of China under Grant 62276222, 61673328, and the Collaborative Project Foundation of Fuzhou-Xiamen-Quanzhou Innovation Demonstration Zone under Grant 3502ZCQXT202001.
\bibliographystyle{IEEEtran}
\bibliography{LMOMCTS}

\begin{thebibliography}{10}
\providecommand{\url}[1]{#1}
\csname url@samestyle\endcsname
\providecommand{\newblock}{\relax}
\providecommand{\bibinfo}[2]{#2}
\providecommand{\BIBentrySTDinterwordspacing}{\spaceskip=0pt\relax}
\providecommand{\BIBentryALTinterwordstretchfactor}{4}
\providecommand{\BIBentryALTinterwordspacing}{\spaceskip=\fontdimen2\font plus
\BIBentryALTinterwordstretchfactor\fontdimen3\font minus
  \fontdimen4\font\relax}
\providecommand{\BIBforeignlanguage}[2]{{%
\expandafter\ifx\csname l@#1\endcsname\relax
\typeout{** WARNING: IEEEtran.bst: No hyphenation pattern has been}%
\typeout{** loaded for the language `#1'. Using the pattern for}%
\typeout{** the default language instead.}%
\else
\language=\csname l@#1\endcsname
\fi
#2}}
\providecommand{\BIBdecl}{\relax}
\BIBdecl

\bibitem{7155533}
X.~Ma, F.~Liu, Y.~Qi, X.~Wang, L.~Li, L.~Jiao, M.~Yin, and M.~Gong, ``A
  multiobjective evolutionary algorithm based on decision variable analyses for
  multiobjective optimization problems with large-scale variables,'' \emph{IEEE
  Transactions on Evolutionary Computation}, vol.~20, no.~2, pp. 275--298,
  2016.

\bibitem{8681243}
Y.~Tian, X.~Zheng, X.~Zhang, and Y.~Jin, ``Efficient large-scale multiobjective
  optimization based on a competitive swarm optimizer,'' \emph{IEEE
  Transactions on Cybernetics}, vol.~50, no.~8, pp. 3696--3708, 2020.

\bibitem{8315121}
P.~Dai, K.~Liu, L.~Feng, H.~Zhang, V.~C.~S. Lee, S.~H. Son, and X.~Wu,
  ``Temporal information services in large-scale vehicular networks through
  evolutionary multi-objective optimization,'' \emph{IEEE Transactions on
  Intelligent Transportation Systems}, vol.~20, no.~1, pp. 218--231, 2019.

\bibitem{7533424}
K.~Tang, J.~Wang, X.~Li, and X.~Yao, ``A scalable approach to capacitated arc
  routing problems based on hierarchical decomposition,'' \emph{IEEE
  Transactions on Cybernetics}, vol.~47, no.~11, pp. 3928--3940, 2017.

\bibitem{8482477}
Y.~Tian, S.~Yang, L.~Zhang, F.~Duan, and X.~Zhang, ``A surrogate-assisted
  multiobjective evolutionary algorithm for large-scale task-oriented pattern
  mining,'' \emph{IEEE Transactions on Emerging Topics in Computational
  Intelligence}, vol.~3, no.~2, pp. 106--116, 2019.

\bibitem{10.1007/978-3-319-16549-3_56}
D.~Kimovski, J.~Ortega, A.~Ortiz, and R.~Ba{\~{n}}os, ``Parallel cooperation
  for large-scale multiobjective optimization on feature selection problems,''
  in \emph{Applications of Evolutionary Computation}, A.~M. Mora and
  G.~Squillero, Eds.\hskip 1em plus 0.5em minus 0.4em\relax Cham: Springer
  International Publishing, 2015, pp. 693--705.

\bibitem{8781874}
Y.~Zhou, G.~G. Yen, and Z.~Yi, ``A knee-guided evolutionary algorithm for
  compressing deep neural networks,'' \emph{IEEE Transactions on Cybernetics},
  vol.~51, no.~3, pp. 1626--1638, 2021.

\bibitem{9804338}
K.~Zhang, C.~Shen, and G.~G. Yen, ``Multipopulation-based differential
  evolution for large-scale many-objective optimization,'' \emph{IEEE
  Transactions on Cybernetics}, pp. 1--13, 2022.

\bibitem{iqbal2015wireless}
M.~Iqbal, M.~Naeem, A.~Anpalagan, A.~Ahmed, and M.~Azam, ``Wireless sensor
  network optimization: Multi-objective paradigm,'' \emph{Sensors}, vol.~15,
  no.~7, pp. 17\,572--17\,620, 2015.

\bibitem{RN105}
H.~Wang, L.~Jiao, R.~Shang, S.~He, and F.~Liu, ``A memetic optimization
  strategy based on dimension reduction in decision space,'' \emph{Evolutionary
  Computation}, vol.~23, no.~1, 2015.

\bibitem{5415586}
J.~J. Durillo, A.~J. Nebro, C.~A.~C. Coello, J.~Garcia-Nieto, F.~Luna, and
  E.~Alba, ``A study of multiobjective metaheuristics when solving parameter
  scalable problems,'' \emph{IEEE Transactions on Evolutionary Computation},
  vol.~14, no.~4, pp. 618--635, 2010.

\bibitem{6557903}
L.~M. Antonio and C.~A.~C. Coello, ``Use of cooperative coevolution for solving
  large scale multiobjective optimization problems,'' in \emph{2013 IEEE
  Congress on Evolutionary Computation}, 2013, pp. 2758--2765.

\bibitem{RN89}
H.~Zille, H.~Ishibuchi, S.~Mostaghim, and Y.~Nojima, ``A framework for
  large-scale multiobjective optimization based on problem transformation,''
  \emph{IEEE Transactions on Evolutionary Computation}, vol.~22, no.~2, pp.
  260--275, 2018.

\bibitem{CHEN2020457}
H.~Chen, R.~Cheng, J.~Wen, H.~Li, and J.~Weng, ``Solving large-scale
  many-objective optimization problems by covariance matrix adaptation
  evolution strategy with scalable small subpopulations,'' \emph{Information
  Sciences}, vol. 509, pp. 457--469, 2020.

\bibitem{8720021}
Y.~Tian, X.~Zhang, C.~Wang, and Y.~Jin, ``An evolutionary algorithm for
  large-scale sparse multiobjective optimization problems,'' \emph{IEEE
  Transactions on Evolutionary Computation}, vol.~24, no.~2, pp. 380--393,
  2020.

\bibitem{9047876}
Y.~Tian, C.~Lu, X.~Zhang, K.~C. Tan, and Y.~Jin, ``Solving large-scale
  multiobjective optimization problems with sparse optimal solutions via
  unsupervised neural networks,'' \emph{IEEE Transactions on Cybernetics},
  vol.~51, no.~6, pp. 3115--3128, 2021.

\bibitem{doi:10.1080/10286608.2012.672412}
Z.~Srdjevic, M.~Samardzic, and B.~Srdjevic, ``Robustness of ahp in selecting
  wastewater treatment method for the coloured metal industry: Serbian case
  study,'' \emph{Civil Engineering and Environmental Systems}, vol.~29, no.~2,
  pp. 147--161, 2012.

\bibitem{4594481}
M.~A. Salido, F.~Barber, and L.~Ingolotti, ``Robustness in railway
  transportation scheduling,'' in \emph{2008 7th World Congress on Intelligent
  Control and Automation}, 2008, pp. 2880--2885.

\bibitem{ARORA2012491}
J.~S. Arora, ``Chapter 12 - numerical methods for constrained optimum design,''
  in \emph{Introduction to Optimum Design}, 3rd~ed., J.~S. Arora, Ed.\hskip 1em
  plus 0.5em minus 0.4em\relax Boston: Academic Press, 2012, pp. 491--531.

\bibitem{ARORA2012575}
{Jasbir S. Arora}, ``Chapter 14 - practical applications of optimization,'' in
  \emph{Introduction to Optimum Design}, 3rd~ed., J.~S. Arora, Ed.\hskip 1em
  plus 0.5em minus 0.4em\relax Boston: Academic Press, 2012, pp. 575--617.

\bibitem{RN98}
E.~Zitzler, L.~Thiele, M.~Laumanns, C.~M. Fonseca, and V.~G.~d. Fonseca,
  ``Performance assessment of multiobjective optimizers: an analysis and
  review,'' \emph{IEEE Transactions on Evolutionary Computation}, vol.~7,
  no.~2, pp. 117--132, 2003.

\bibitem{RN220}
L.~Kocsis and C.~Szepesv{\'a}ri, ``Bandit based monte-carlo planning,'' in
  \emph{Machine Learning: ECML 2006}, J.~F{\"u}rnkranz, T.~Scheffer, and
  M.~Spiliopoulou, Eds.\hskip 1em plus 0.5em minus 0.4em\relax Berlin,
  Heidelberg: Springer Berlin Heidelberg, 2006, pp. 282--293.

\bibitem{RN57}
R.~Y. Rubinstein and D.~P. Kroese, \emph{Simulation and the Monte Carlo
  Method}.\hskip 1em plus 0.5em minus 0.4em\relax John Wiley and Sons, 2008.

\bibitem{HAZRA2019454}
A.~Hazra, V.~Maggioni, P.~Houser, H.~Antil, and M.~Noonan, ``A monte
  carlo-based multi-objective optimization approach to merge different
  precipitation estimates for land surface modeling,'' \emph{Journal of
  Hydrology}, vol. 570, pp. 454--462, 2019.

\bibitem{BARNOON20222747}
P.~Barnoon, D.~Toghraie, B.~Mehmandoust, M.~A. Fazilati, and S.~A. Eftekhari,
  ``Natural-forced cooling and monte-carlo multi-objective optimization of
  mechanical and thermal characteristics of a bipolar plate for use in a proton
  exchange membrane fuel cell,'' \emph{Energy Reports}, vol.~8, pp. 2747--2761,
  2022.

\bibitem{6145622}
C.~B. Browne, E.~Powley, D.~Whitehouse, S.~M. Lucas, P.~I. Cowling,
  P.~Rohlfshagen, S.~Tavener, D.~Perez, S.~Samothrakis, and S.~Colton, ``A
  survey of monte carlo tree search methods,'' \emph{IEEE Transactions on
  Computational Intelligence and AI in Games}, vol.~4, no.~1, pp. 1--43, 2012.

\bibitem{fasthv}
J.~Bader, K.~Deb, and E.~Zitzler, ``Faster hypervolume-based search using monte
  carlo sampling,'' in \emph{Multiple Criteria Decision Making for Sustainable
  Energy and Transportation Systems}, M.~Ehrgott, B.~Naujoks, T.~J. Stewart,
  and J.~Wallenius, Eds.\hskip 1em plus 0.5em minus 0.4em\relax Berlin,
  Heidelberg: Springer Berlin Heidelberg, 2010, pp. 313--326.

\bibitem{10.5555/1762545.1762618}
E.~Zitzler, D.~Brockhoff, and L.~Thiele, ``The hypervolume indicator revisited:
  On the design of pareto-compliant indicators via weighted integration,'' in
  \emph{Proceedings of the 4th International Conference on Evolutionary
  Multi-Criterion Optimization}, ser. EMO'07.\hskip 1em plus 0.5em minus
  0.4em\relax Berlin, Heidelberg: Springer-Verlag, 2007, p. 862–876.

\bibitem{tian2021evolutionary}
Y.~Tian, L.~Si, X.~Zhang, R.~Cheng, C.~He, K.~C. Tan, and Y.~Jin,
  ``Evolutionary large-scale multi-objective optimization: A survey,''
  \emph{ACM Computing Surveys}, vol.~1, no.~1, 2021.

\bibitem{RN82}
X.~Zhang, Y.~Tian, R.~Cheng, and Y.~Jin, ``A decision variable clustering-based
  evolutionary algorithm for large-scale many-objective optimization,''
  \emph{IEEE Transactions on Evolutionary Computation}, vol.~22, no.~1, pp.
  97--112, 2018.

\bibitem{RN259}
J.~Lin, C.~He, and R.~Cheng, ``Adaptive dropout for high-dimensional expensive
  multiobjective optimization,'' \emph{Complex \& Intelligent Systems}, vol.~8,
  no.~1, pp. 271--285, 2022.

\bibitem{CAO2020100626}
B.~Cao, J.~Zhao, Y.~Gu, Y.~Ling, and X.~Ma, ``Applying graph-based differential
  grouping for multiobjective large-scale optimization,'' \emph{Swarm and
  Evolutionary Computation}, vol.~53, p. 100626, 2020.

\bibitem{9723458}
S.~Liu, J.~Li, Q.~Lin, Y.~Tian, and K.~C. Tan, ``Learning to accelerate
  evolutionary search for large-scale multiobjective optimization,'' \emph{IEEE
  Transactions on Evolutionary Computation}, pp. 1--1, 2022.

\bibitem{RN90}
C.~He, S.~Huang, R.~Cheng, K.~C. Tan, and Y.~Jin, ``Evolutionary multiobjective
  optimization driven by generative adversarial networks (gans),'' \emph{IEEE
  Transactions on Cybernetics}, pp. 1--14, 2020.

\bibitem{QI20221601}
S.~Qi, J.~Zou, S.~Yang, Y.~Jin, J.~Zheng, and X.~Yang, ``A self-exploratory
  competitive swarm optimization algorithm for large-scale multiobjective
  optimization,'' \emph{Information Sciences}, vol. 609, pp. 1601--1620, 2022.

\bibitem{GE20221441}
Y.~Ge, D.~Chen, F.~Zou, M.~Fu, and F.~Ge, ``Large-scale multiobjective
  optimization with adaptive competitive swarm optimizer and inverse
  modeling,'' \emph{Information Sciences}, vol. 608, pp. 1441--1463, 2022.

\bibitem{RN257}
H.~Hong, K.~Ye, M.~Jiang, D.~Cao, and K.~C. Tan, ``Solving large-scale
  multiobjective optimization via the probabilistic prediction model,''
  \emph{Memetic Computing}, 2022.

\bibitem{1197687}
E.~Zitzler, L.~Thiele, M.~Laumanns, C.~Fonseca, and V.~da~Fonseca,
  ``Performance assessment of multiobjective optimizers: an analysis and
  review,'' \emph{IEEE Transactions on Evolutionary Computation}, vol.~7,
  no.~2, pp. 117--132, 2003.

\bibitem{10.1007/3-540-36970-8_37}
M.~Fleischer, ``The measure of pareto optima applications to multi-objective
  metaheuristics,'' in \emph{Evolutionary Multi-Criterion Optimization}, C.~M.
  Fonseca, P.~J. Fleming, E.~Zitzler, L.~Thiele, and K.~Deb, Eds.\hskip 1em
  plus 0.5em minus 0.4em\relax Berlin, Heidelberg: Springer Berlin Heidelberg,
  2003, pp. 519--533.

\bibitem{996017}
K.~Deb, A.~Pratap, S.~Agarwal, and T.~Meyarivan, ``A fast and elitist
  multiobjective genetic algorithm: Nsga-ii,'' \emph{IEEE Transactions on
  Evolutionary Computation}, vol.~6, no.~2, pp. 182--197, 2002.

\bibitem{7553457}
R.~Cheng, Y.~Jin, M.~Olhofer, and B.~Sendhoff, ``Test problems for large-scale
  multiobjective and many-objective optimization,'' \emph{IEEE Transactions on
  Cybernetics}, vol.~47, no.~12, pp. 4108--4121, 2017.

\bibitem{Haynes2013}
W.~Haynes, \emph{Wilcoxon Rank Sum Test}.\hskip 1em plus 0.5em minus
  0.4em\relax New York, NY: Springer New York, 2013, pp. 2354--2355.

\bibitem{1583625}
L.~While, P.~Hingston, L.~Barone, and S.~Huband, ``A faster algorithm for
  calculating hypervolume,'' \emph{IEEE Transactions on Evolutionary
  Computation}, vol.~10, no.~1, pp. 29--38, 2006.

\bibitem{RN92}
Y.~Tian, R.~Cheng, X.~Zhang, and Y.~Jin, ``Platemo: A matlab platform for
  evolutionary multi-objective optimization [educational forum],'' \emph{IEEE
  Computational Intelligence Magazine}, vol.~12, no.~4, pp. 73--87, 2017.

\bibitem{RN88}
C.~He, L.~Li, Y.~Tian, X.~Zhang, R.~Cheng, Y.~Jin, and X.~Yao, ``Accelerating
  large-scale multiobjective optimization via problem reformulation,''
  \emph{IEEE Transactions on Evolutionary Computation}, vol.~23, no.~6, pp.
  949--961, 2019.

\bibitem{20.500.11850/84395}
J.~Knowles, L.~Thiele, and E.~Zitzler, ``\BIBforeignlanguage{en}{A tutorial on
  the performance assessment of stochastic multiobjective optimizers},''
  Guanajuato, 2005, third International Conference on Evolutionary
  Multi-Criterion Optimization (EMO 2005); Conference Date: March 9-11, 2005;
  Conference lecture.

\bibitem{1705400}
S.~Huband, P.~Hingston, L.~Barone, and L.~While, ``A review of multiobjective
  test problems and a scalable test problem toolkit,'' \emph{IEEE Transactions
  on Evolutionary Computation}, vol.~10, no.~5, pp. 477--506, 2006.

\bibitem{8962275}
C.~He, R.~Cheng, C.~Zhang, Y.~Tian, Q.~Chen, and X.~Yao, ``Evolutionary
  large-scale multiobjective optimization for ratio error estimation of voltage
  transformers,'' \emph{IEEE Transactions on Evolutionary Computation},
  vol.~24, no.~5, pp. 868--881, 2020.

\bibitem{9097186}
M.~Jiang, Z.~Wang, L.~Qiu, S.~Guo, X.~Gao, and K.~C. Tan, ``A fast dynamic
  evolutionary multiobjective algorithm via manifold transfer learning,''
  \emph{IEEE Transactions on Cybernetics}, vol.~51, no.~7, pp. 3417--3428,
  2021.

\bibitem{9199822}
M.~Jiang, Z.~Wang, S.~Guo, X.~Gao, and K.~C. Tan, ``Individual-based transfer
  learning for dynamic multiobjective optimization,'' \emph{IEEE Transactions
  on Cybernetics}, vol.~51, no.~10, pp. 4968--4981, 2021.

\end{thebibliography}

\newpage

 




\vfill

\end{document}